\documentclass[pdflatex,sn-basic]{sn-jnl}

\usepackage[english]{babel}

\usepackage{float}
\usepackage{booktabs}
\usepackage{amsmath}
\usepackage{amsthm}
\usepackage{graphicx}
\usepackage{balance} 
\usepackage{needspace} 
\usepackage{booktabs}
\usepackage{algorithm}
\usepackage{algpseudocode}
\usepackage{diagbox}
\usepackage{tabularx}
\usepackage{verbatim}
\usepackage{balance}
\usepackage{caption}

\jyear{2022}

\newtheorem{definition}{Definition}

\begin{document}

\title[ClaSP - Parameter-free Time Series Segmentation]{ClaSP - Parameter-free Time Series Segmentation}

\author*[1]{\fnm{Arik} \sur{Ermshaus}}\email{ermshaua@informatik.hu-berlin.de}

\author*[1]{\fnm{Patrick} \sur{Sch\"afer}}\email{patrick.schaefer@hu-berlin.de}

\author*[1]{\fnm{Ulf} \sur{Leser}}\email{leser@informatik.hu-berlin.de}

\affil[1]{\orgname{Humboldt Universit\"at zu Berlin}, \state{Berlin}, \country{Germany}}

\abstract{
The study of natural and human-made processes often results in long sequences of temporally-ordered values, aka time series (TS). Such processes often consist of multiple states, e.g. operating modes of a machine, such that state changes in the observed processes result in changes in the distribution of shape of the measured values. Time series segmentation (TSS) tries to find such changes in TS post-hoc to deduce changes in the data-generating process. TSS is typically approached as an unsupervised learning problem aiming at the identification of segments distinguishable by some statistical property. Current algorithms for TSS require domain-dependent hyper-parameters to be set by the user, make assumptions about the TS value distribution or the types of detectable changes which limits their applicability. Common hyper-parameters are the measure of segment homogeneity and the number of change points, which are particularly hard to tune for each data set. We present ClaSP, a novel, highly accurate, hyper-parameter-free and domain-agnostic method for TSS. ClaSP hierarchically splits a TS into two parts. A change point is determined by training a binary TS classifier for each possible split point and selecting the one split that is best at identifying subsequences to be from either of the partitions. ClaSP learns its main two model-parameters from the data using two novel bespoke algorithms. In our experimental evaluation using a benchmark of 107 data sets, we show that ClaSP outperforms the state of the art in terms of accuracy and is fast and scalable. Furthermore, we highlight properties of ClaSP using several real-world case studies.}


\sloppy
\maketitle

\section{Introduction}

Recent years brought an explosion in applications for low-cost high resolution sensors, for instance in mobile devices, manufacturing monitoring, or environmental and medical surveillance \citep{Carvalho2019ASL}. These sensors produce large amounts of unlabeled temporally-ordered, real-valued sequences, also referred to as data series or \emph{time series (TS)}. Such signals capture the inherent statistical properties and temporal patterns of processes. Abrupt shifts or \emph{change points (CPs)} of these properties in TS indicate state transitions in the data-generating process. In human activity analysis (HAR), for instance, sensor data from accelerometers captures human behavior. Shifts in rotational deflections indicate changes in activity types that can be used to gather insight on health status or provide activity-aware services \citep{Feuz2015AutomatedDO}. Another example is medical condition monitoring, where a patient's physiological variables like heart rate or electroencephalogram (EEG) are continuously monitored and analyzed to detect trends and anomalies. Clinicians use change points in the corresponding signals to identify epilepsy or sleep problems~\citep{Malladi2013OnlineBC}. The semantic segmentation of medical images can further detect cardiac substructures that may reveal cardiovascular conditions, or locate the position and shape of constituent bones from which patient age can be estimated~\citep{Janik2021InterpretabilityOA,Davis2012OnTS}. In predictive maintenance applications, sensors monitor industrial machinery and may detect their malfunctioning as changes in the measurements. The early identification of wear and tear in sensor recordings is crucial to ensure machine availability \citep{Zenisek2019MachineLB}. The importance of these applications leads to an increasing interest in the problem of change point detection (CPD)~\citep{truong2020selective} and time series segmentation (TSS)~\citep{aminikhanghahi2017survey}. These two problems are complementary: While CPD aims at finding the exact positions of change points in a TS, TSS aims at partitioning a TS into disjoint segments corresponding to states of the data-generating process. Solving CPD thus naturally leads to a TSS method and vice versa, as change points exactly depict the indices of a TS where segments change. 

A variety of TSS algorithms have been proposed, see~\citep{aminikhanghahi2017survey} for a survey. Most of them are domain-specific ~\citep{brahim2004gaussian, bosc2003automatic, Zenisek2019MachineLB}. In contrast, only few domain-agnostic TSS algorithms exist, including FLOSS~\citep{gharghabi2017matrix}, Autoplait~\citep{matsubara2014autoplait}, and HOG-1D~\citep{zhao2016decomposing}. FLOSS is the state-of-the-art domain-agnostic TSS algorithm. Given a subsequence length and the number of change points as hyper-parameters, it first annotates a TS with a bespoke \emph{arc curve}, which spans from each offset in the TS to its 1-NN subsequence. The offsets with the least crossings of arc curves indicate the potential change points. FLOSS outperforms Autoplait and HOG-1D in an experimental evaluation of 32 TS~\citep{gharghabi2017matrix}. However, it requires the user to set two hyper-parameters and degrades in performance for TS that capture processes with repeating states~\citep{Deldari2020ESPRESSOEA}. The two hyper-parameters are domain-dependent, which is why they are pre-defined for many benchmark data sets in the literature.

In this work, we present ClaSP (Classification Score Profile), a new, highly accurate parameter-free and domain-agnostic algorithm for TSS. ClaSP approaches segmentation by reducing the TSS problem to a binary TS classification problem of identifying regions by similar shape. The underlying assumption is that subsequences extracted from the same segment of a TS are \emph{self-similar} (as in mutually similar), but dissimilar from subsequences extracted from other segments. The central idea of ClaSP is to iteratively determine change points by finding the point in the TS where the performance of a binary classifier, trained to separate subsequences either to belong to the left or the right part, is highest. This allows the usage of established methods from supervised TS analysis for solving an unsupervised TS problem. A potential drawback of this approach is the runtime necessary for training and evaluating individual classifiers for many possible hypothetical splits. However, we show that ClaSP, when using a $k$-NN classifier, achieves both highly accurate results and also is very fast, because most of the work can be factored out into a preprocessing phase. Virtually all state-of-the-art methods have domain-dependent hyper-parameters that need to be set by human experts. Other than that, ClaSP learns its (model-)parameters from a data set at hand using self-supervision and two novel bespoke techniques.

\begin{figure}[t]
	\includegraphics[width=1.0\columnwidth]{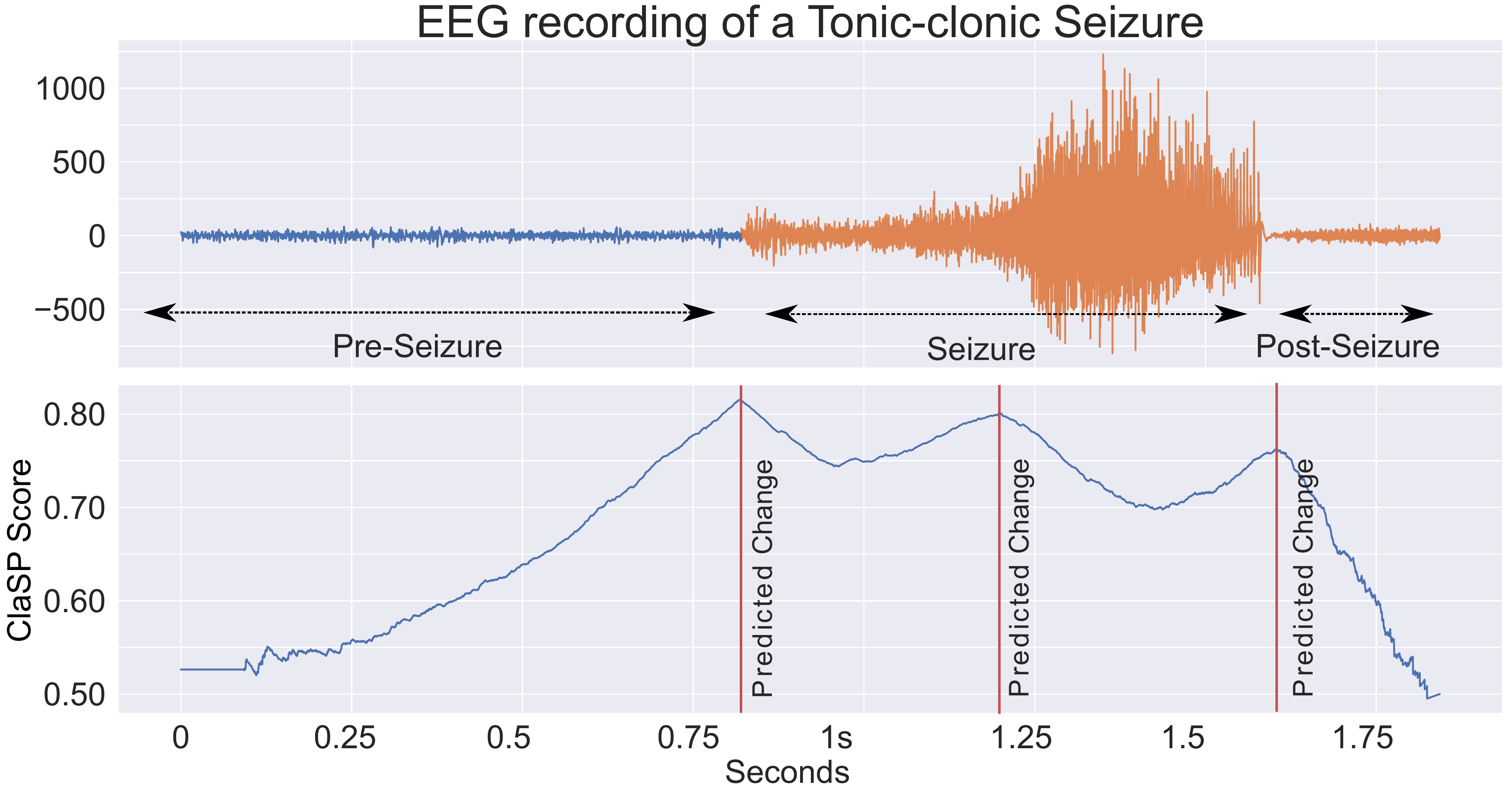}
	\caption{A scalp right central electrode recording of an EEG from a subject showing a tonic-clonic seizure~\citep{quiroga1997searching}. It contains three segments that capture pre-seizure, seizure and post-seizure EEG activity.
	The three local maxima in \emph{ClaSP} (bottom) highlight potential change points based on changes in the underlying shapes of the divided segments, representing the three activities.\label{fig:ClaSP}
	}
\end{figure}

Figure~\ref{fig:ClaSP} illustrates ClaSP together with the corresponding TS. The data set originates from~\citep{quiroga1997searching} and was preprocessed by~\citep{gharghabi2017matrix}. It shows an EEG of a human brain with tonic-clonic seizure activity after about 1 minute of recordings. ClaSP first creates a profile (called \emph{Classification Score Profile} or \emph{ClaSP}) which scores the accuracy of the hypothetical classifiers at selected hypothetical split points. Based on the ClaSP, change points can easily be identified using a bespoke peak detection algorithm in combination with a non-parametric significance test. In the figure, the three local maxima in ClaSP (bottom) mark the change points between dissimilar segments. The first segment captures the pre-seizure, the two following the seizure (made of two distinct phases), and the last segment corresponds to post-seizure activities. The global maximum in the profile is between the first and second segment (first red vertical line) and matches the start of the seizure. 

ClaSP and its application to TSS is the main focus of this paper. However, we make a special effort to foster reproducible and comparable TSS research to enable follow-up works. The specific contributions of this paper are:

\begin{enumerate}
    \item We present \emph{Classification Score Profile} (ClaSP), a novel (hyper-)parameter-free, domain-agnostic method for TSS using the concept of self-supervision~\citep{tsai2020demystifying}. The profile computed by ClaSP also allows for intuitive visualizations, making segmentation decisions intuitive to communicate and easy to interpret for humans.
    
    \item We present \emph{Summary Statistics Subsequence} (SuSS), a novel window size selection algorithm that determines those window sizes which share similar statistical properties with the entire TS. We use SuSS to learn the length of the subsequences in ClaSP.

    \item We present a novel change point validation test, that uses non-parametric hypothesis testing to directly report statistically significant CPs. This test enables ClaSP to learn the number of CPs automatically.

    \item We provide a Python implementation of ClaSP which achieves a runtime as fast as competitor methods~\citep{ClaSPWebpage} with a segmentation accuracy unmatched by all competitors.
    
    \item We performed a series of experiments on $107$ TS from \citep{TSSBWebpage,gharghabi2017matrix} and compare results to six state-of-the-art competitors, namely FLOSS, BinSeg, PELT, Window, BOCD, and ESPRESSO~\citep{gharghabi2017matrix,truong2020selective,adams2007bayesian,Deldari2020ESPRESSOEA}. To date this is the largest TSS benchmark. ClaSP exhibits the highest overall segmentation performance and achieves the best segmentations in $70$ out of the $107$ cases.
    
    \item We release the Time Series Segmentation Benchmark (TSSB)~\citep{TSSBWebpage}, consisting of $75$ TS annotated with CPs and window sizes based on TS from the UEA/UCR  archive~\citep{UCRClassification} to foster future research.
\end{enumerate}

We first introduced ClaSP in \citep{Schfer2021ClaSPT}. This paper extends ClaSP by making it parameter-free, introducing for the first time SuSS (Section~\ref{sec:window-size-selec}), the change point validation test (Section~\ref{sec:cp-candidate-val}), and TSS with reoccurring segments (Section~\ref{sec:ensembling}).

The remainder of this work is organized as follows: Section~\ref{sec:background} presents background and definitions. Section~\ref{sec:related_work} introduces related work. Section~\ref{sec:clasp_method} presents the ClaSP method. Section~\ref{sec:experiments} presents our experimental evaluation, and Section~\ref{sec:conclusion} concludes the paper.

\section{Background and Definitions}\label{sec:background}

In this work, we assume that a TS is generated by observing some output of an underlying process. This process has several distinct states and switches between these states at arbitrary and a priori unknown points in time. We furthermore assume that different states produce (slightly) different measurements~\citep{Gharghabi2018DomainAO}; finding these points in time when the process changes state by analyzing the different parts of a TS is the task of change point detection and also leads to a segmentation of a TS. We now define these concepts formally.

\begin{definition}
A process $P = (S,T)$ consists of one or more discrete states $s_1, ..., s_l \in S$ that are pairwise separated by transitions $(s_i,s_k) \in T \subseteq  S \times S$ with $s_i \neq s_k$.
\end{definition}

\begin{definition}
A \emph{time series (TS)} $T$ is a sequence of $n \in \mathbb{N}$ real values, $T=(t_{1},\ldots,t_{n}),t_{i} \in \mathbb{R}$ that measures an observable output of a process $P$. The values are also called data points.
\end{definition}

\begin{definition}
Given a TS $T$, a \emph{subsequence} $T_{s,e}$ of $T$ with start offset $s$ and end offset $e$ consists of the contiguous values of $T$ from position $s$ to position $e$, i.e., $T_{s,e}=(t_{s},\text{\dots},t_{e})$ with $1\leq s \leq e \leq n$. The length of $T_{s,e}$ is $\|T_{s,e}\| = e-s+1$.
\end{definition}

We sometimes call subsequences \emph{windows} and their length \emph{width}.

\begin{definition}
A \emph{periodic} TS is one that approximately repeats a subsequence of values after a fixed length of time, known as the period.
\end{definition}

We only consider periodic TS in this paper. However, local parts of a TS can nevertheless deviate from each other, e.g. in period length, shape or amplitude. 

\begin{definition}
Given a process $P$ and a corresponding TS $T$, a \emph{change point (CP, or split)} is an offset $i \in [1,\dots,n]$ that corresponds to a state transition in $P$. The problem of change point detection (CPD) is to identify all CPs / splits in $T$.
\end{definition}

CPD is often approached by first fixing the number $C$ of segments; the task then becomes finding $C-1$ splits (change points) in a TS. In such a setting, $C$ is a hyper-parameter that is as difficult to infer from the data as in finding the correct number of clusters in clustering algorithms~\citep{Nguyen2015}. Note that ClaSP has a build-in method to find the optimal $C$ of a TS according to a statistical significance test (see Subsection~\ref{sec:cp-candidate-val}). Segmentation, i.e., finding subsequences of a TS produced by the same state of the underlying process, is directly related to CPD and defined as follows.

\begin{definition}
Given a process $P$ and a corresponding TS $T$, a \emph{segmentation} of $T$ is the ordered sequence of indices of $T$, i.e., $t_{i_{1}}$,...,$t_{i_{S}}$ with ${1<i_1<\dots<i_S<n}$ at which the underlying process $P$ changed its state. 
\end{definition}

Naturally, an algorithm which produced $C-1$ change points induces a segmentation of $C$ segments and vice versa. Note that we make no assumptions regarding the frequency of occurrences of states, i.e., a process may start from a state $A$, then shift to $B$, back to $A$ etc.

\subsection{Time Series Classification}

ClaSP, the algorithm we propose in this paper, is based on  \emph{time series classification (TSC)} and self-supervision. I.e., it solves the unsupervised TSS problem through a reduction to a binary TS classification problem and using a classifier to derive change points. Here, we formally define the concepts used in TSC. 

\begin{definition}
A binary TSC data set $D = (X, Y)$, $X \subset \mathbb{R}^{m \times n}$, $Y = \{0,1\}^m$ consists of $m$ TS $T \in X$, $\|T\| = n$ annotated with associated labels $y \in Y$. 
\end{definition}

We use the terms \emph{labels} and \emph{classes} interchangeably. Note that in classical TSC, a data set is annotated with labels by human experts for training. In contrast, ClaSP learns by automatically labelling subsequences around hypothetical split points using the concept of self-supervision.

\begin{definition}
A time series classification algorithm $\textsc{clf}$ maps an unlabeled TS $T$ to a label $y$ according to a model that was learned from a labelled training data set $D = (X, Y)$. The problem of time series classification (TSC) is to map $T$ to the correct label $y$ using $\textsc{clf}$ trained on $D$.
\end{definition}

Many $\textsc{clf}$ are based on some notion of shape-based similarity between TS. The literature differentiates distance and feature-based approaches~\citep{bagnall2016great}. The former compare distances between whole time series while the latter extract statistical properties or temporal patterns as features from TS. ClaSP implements a distance-based classifier (see Subsection~\ref{sec:classifier}), but we also discuss other variants in the experimental evaluation (see Subsection~\ref{sec:complex-classifiers}).

\subsection{Binary Classification Evaluation Metrics}
\label{sec:eval_metrics}

We evaluate the TSC problem for ClaSP with a binary classification \emph{evaluation metric}, referred to as score in the remainder of this paper. The score is a function that maps a sequence of $m$ ground truth class labels $Y_t = \{0,1\}^m$ in conjunction with a sequence of $m$ predicted labels $Y_p = \{0,1\}^m$ to a normalized real value $c \in [0,\dots,1]$. It assesses the prediction quality of a machine learning algorithm. Common scores are defined on the basis of true positives (TP), false positives (FP), false negatives (FN) and true negatives (TN). Based on this categorization, we use the following scores for designing and evaluating ClaSP:
\begin{align}
    \text{Precision} &= \frac{\|TP\|}{\|TP\| + \|FP\|} \text{, }
    \text{Recall} = \frac{\|TP\|}{\|TP\| + \|FN\|} \\
    \text{F}_1 &= 2 \cdot \frac{\text{Precision } \cdot \text{ Recall}}{\text{Precision + Recall}} \\
    \text{TPR} &= \frac{\|TP\|}{\|TP\| + \|FN\|}, 
    \text{FPR} = \frac{\|FP\|}{\|FP\| + \|TN\|} \\
    \text{ROC/AUC} &= \int_{x=0}^{1} \mbox{TPR}(\mbox{FPR}^{-1}(x))\text{ } dx 
\end{align}

We introduce and use different metrics because the problems ClaSP classifies vary greatly in terms of class imbalance. At the same time, different metrics are affected in different ways by class imbalance. The F1 score, defined as the harmonic mean between Precision and Recall, is a good choice for scenarios with a small fraction of positive samples. ROC/AUC~\citep{fawcett2006introduction}, on the other hand, is known to be insensitive to class-imbalance. The ROC curve plots the true positive rate (TPR) against the false positive rate (FPR) for varying thresholds. The AUC score summarizes the curve as the probability that the classifier, at hand, is able to tell positive and negative samples apart. 

Evaluation scores further can be distinguished into \emph{micro} and \emph{macro} averages. Micro averages are computed globally by counting the total TP and FN as well as FP and FN, for $y=0$ and $y=1$. Opposed to that, macro averages calculate these groups for each class label separately for both classes to compute their unweighted mean. For ClaSP, we only apply macro averaged scores (see Subsection~\ref{sec:scoring}).

\section{Related Work}\label{sec:related_work}

A number of domain-specific TSS methods that can detect a limited range of signal changes in TS with suitable value distributions (e.g. piecewise-constant or Gaussian) have been published in the last decades. Truong et al.~\citep{truong2020selective} present a review of such techniques. They compare methods regarding their cost function, search method, and whether the number of change points is known a priori. They discern three main classes: (a) likelihood-based methods, (b) kernel-based methods, and (c) graph-based methods. Likelihood-based methods split TS into consecutive windows and compare the probability distributions of windows~\citep{kawahara2012sequential}. If these differ significantly, a change point is introduced. Kernel-based methods also split the TS into windows and then use a kernel-based statistical test to assess the homogeneity between subsequent windows~\citep{harchaoui2009regularized}. Graph-based methods first infer a graph by mapping observations (i.e. windows or sets of TS) to nodes and connecting nodes by edges if their pairwise similarity exceeds a predefined threshold. Next, a bespoke graph statistic is applied to split the graph into sub-graphs leading to change points in the TS~\citep{chen2015graph}. Katser et al.~\citep{Katser2021UnsupervisedOC} analyses and formalizes ensembling strategies to overcome domain-specific model selection and increase robustness and accuracy. 

Search methods like Binary Segmentation (BinSeg)~\citep{Sen1975OnTF}, Pruned Exact Linear Time (PELT)~\citep{Killick2012OptimalDO}, or window-based segmentation (Window)~\citep{truong2020selective} use the aforementioned approaches to find meaningful segmentations that minimize the associated costs. Another segmentation approach is Bayesian Online Change Point Detection (BOCD)~\citep{adams2007bayesian}, that calculates the probability distribution of data points since the last change using a recursive message-passing algorithm to infer the most recent change point. Draayer et al.~\citep{Draayer2021ReevaluatingTC} extend this idea for short gradual changes. A recent benchmark compared the aforementioned TSS methods on a wide variety of small to medium-sized, non-periodic, univariate and multivariate TS~\citep{van2020evaluation} and found that BinSeg, PELT and BOCD outperform other domain-specific techniques. 

All the aforementioned methods require a user to set domain-specific model-parameters or assume a TS to follow a specific value distribution. Domain-agnostic segmentation solutions, that impose no assumptions on the observed data points and change types, have seen much less popularity. The most prominent of these algorithms is FLOSS~\citep{gharghabi2017matrix,Gharghabi2018DomainAO}. It uses the proximity of a window to the most similar other window to create an arc curve, which is a vector that contains for each index $i$ the number of \emph{arcs} that cross over $i$. Local minima of this number indicate boundaries (change points) of homogenous regions. Deldari et al.~\citep{Deldari2020ESPRESSOEA} published ESPRESSO, a variant of FLOSS that extends the arc curve with TS chains~\citep{Zhu2017MatrixPV} and weighs it with positional subsequence information in order to detect reoccurring segments. It also uses a more sophisticated entropy-based segmentation procedure. 

We shall compare ClaSP to the latter $6$ competitors regarding accuracy and runtimes in Section~\ref{sec:experiments}. The main shortcoming of current domain-agnostic TSS algorithms is that they require a user to set data-dependent hyper-parameters, e.g. the subsequence length or amount of CPs. In contrast, ClaSP can learn its main two parameters automatically, though the experienced domain expert may also set them manually. Furthermore, ClaSP is the first TSS algorithm that uses self-supervision. A similar technique is used in~\citep{hido2008unsupervised}, but in this case applied for detecting differences between two data sets. Note that the problem we study is very different from supervised TSS, where a model is trained on TS with annotated segments~\citep{cook2015activity}; in unsupervised TSS, no such annotation is required or given. 

The quantitative evaluation of segmentation algorithms is difficult, as the semantics in TS can be ambiguous and high quality ground truth annotations are typically scarce. Large fractions of the segmentation research is either only theoretically analysed or experimentally evaluated on synthetic data sets~\citep{truong2020selective}. While this is very useful for testing specific scenarios, it has limited meaning for real-world applications. To tackle these shortcomings, it is common practice in the unsupervised time series data mining literature to evaluate algorithms on semi-synthetic data sets, e.g. ones with real-world data that is in part (re-)arranged or changed to create genuine problem settings and inferred ground truths labels. For anomaly detection, the largest benchmark contains $250$ TS from natural, human and animal processes with synthetic anomalies injected at arbitrary locations~\citep{KeoghMultiDatasetTADC}. Similarly, for segmentation, Gharghabi et al. also published a benchmark that contains data from $8$ real as well as $7$ semi-synthethic and one synthetic use case(s), which are e.g. concatenated or flipped recordings of different samples~\citep{gharghabi2017matrix}. In the same fashion, we release more semi-synthetic annotated TS to introduce new challenging use cases from various sensor types with diverse problem settings~\citep{TSSBWebpage}. In Section~\ref{sec:experiments}, we shall evaluate ClaSP and its competitors on the existing TS and our new data sets.

\section{ClaSP - Classification Score Profile}\label{sec:clasp_method}

\begin{figure}[t]
    \centering
	\includegraphics[width=0.8\columnwidth]{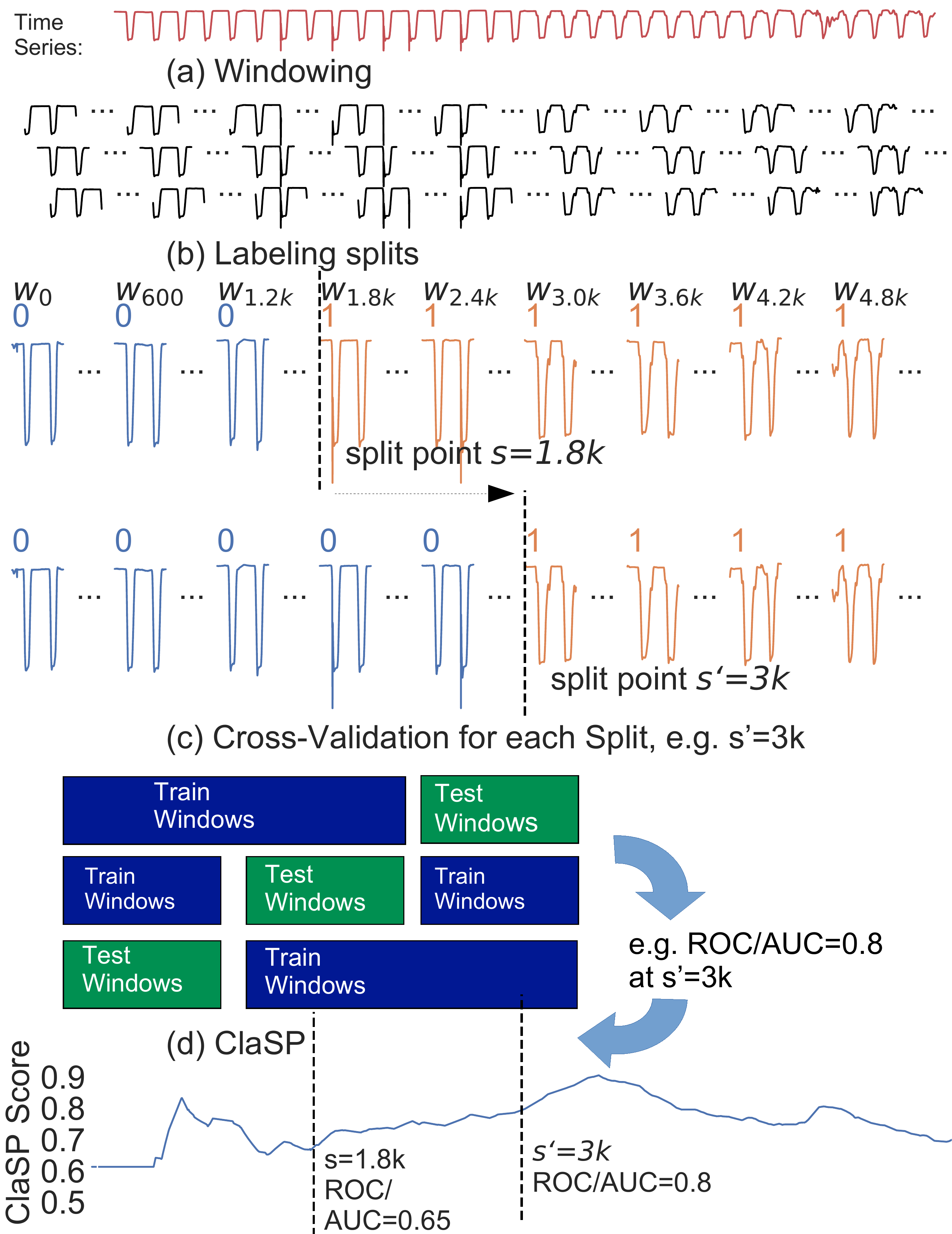}
	\caption{(a) A TS (in red) is split into overlapping windows (in black). (b) These windows are labelled with $0$ for windows to the left and $1$ to the right of hypothetical split points, exemplary depicted for indexes $1800$ and $3000$. (c) For each hypothetical split, a binary classifier is trained and evaluated using cross-validation. (d) Finally, the ClaSP is created from the cross-validation score of the classifiers.
	\label{fig:clasp_labeling}
	}
\end{figure}

\emph{Classification Score Profile (ClaSP)} is a novel method for TS segmentation based on self-supervision. We first give an intuitive overview of our method, before we explain its individual steps in detail in Sections~\ref{sec:classifier} to~\ref{sec:model-param}. We primarily introduce ClaSP as a method for single CPD and describe extensions necessary for approaching the general TSS problem in Sections~\ref{sec:segmentation_alg} and~\ref{sec:cp-candidate-val}. Section~\ref{sec:complexity} analyses the computational complexity of our methods.

Given a TS $T$ with length $\|T\|=n$, ClaSP first initializes an empty classification score profile. To this end, we partition $T$ into overlapping windows of a fixed length $w$ with similar shape (statistical properties). The optimal subsequence length $w$ is learned using a bespoke algorithm from the data (see Subsection~\ref{sec:window-size-selec}). Next, hypothetical splits are generated for increasing offsets $i \in [w+1,\dots,n-w-1]$. This creates hypothetical segments. For each such segment, certain characteristics are extracted for all of its windows as features for the self-supervision. We transform each such split into a binary classification problem $Y=\{0,1\}$ by attaching label $0$ ($1$) to all windows to the left (right) of the split point. A binary $k$-Nearest-Neighbor classifier ($k$-NN) is trained on these features and evaluated in a cross-evaluation setting; see Subsection~\ref{sec:complex-classifiers} for results with other classifiers. We interpret the cross-validation score of the classifier as a measure of how dissimilar windows from the left hypothetical segment are to windows from the right, where a high score means low similarity. This degree of \emph{intra-segment similarity} (self-similarity) is recorded for each offset $i$, together forming the classification score profile for $T$. Every local maximum in this profile represents a potential change point, as it is a point where the differences between the TS part to the left and that to the right is the highest (compare Figure~\ref{fig:ClaSP}). 

\begin{algorithm}[t]
	\caption{Classification Score Profile}\label{alg:clasp}
	\begin{algorithmic}[1]
		\Procedure{calc\_clasp}{$T$, $w$}		
			
			\State $\textsc{ClaSP} \gets $ initialize array of length $\|T\|$ with 0
			\State $Y_t \gets $ initialize array of length $\|T\|$ with $1$ \Comment{all labels in $W_R$}
			
			\State $W = \textsc{windows}(T, w)$ \Comment{all windows from $T$}
			\State $\textsc{kNNs} = \textsc{knn\_profile}(W)$
	    
			\For{$i \in [w+1 , \dots, \|T\|-w-1) $} \Comment{for each split}
                \State $Y_t[i-w] \gets 0$ \Comment{add current window to $W^L_i$}
			    \State $\textsc{ClaSP}[i] \gets \textsc{cross\_validate}(\textsc{kNNs},Y_t$)
			\EndFor 
			\State \Return{$\textsc{ClaSP}$}
		\EndProcedure
	\end{algorithmic}
\end{algorithm}

The high-level workflow of ClaSP is shown graphically in Figure~\ref{fig:clasp_labeling}. Pseudocode is shown in Algorithm~\ref{alg:clasp}. It first computes $n-w-1$ overlapping windows of width $w$ (line~4, Figure~\ref{fig:clasp_labeling}a) with which it calculates a $k$-NN as preprocessing (line~5). For a given split $i$, ClaSP creates a binary classification problem by labelling the first $[0,\dots,i-w]$ windows $0$ and the remaining $[i-w+1,\dots,n-w-1]$ windows $1$ (line~7, Figure~\ref{fig:clasp_labeling}b). We evaluate the $k$-NN classifier with the labelled windows and store the associated cross-validation score in the classification score profile at offset $i$ (line~8, Figure~\ref{fig:clasp_labeling}c). ClaSP eventually reports the profile (line~10, Figure~\ref{fig:clasp_labeling}d) from which CPs can be easily identified as local maxima.

In the following subsections, we elaborate on (\ref{sec:classifier}) the choice of using a $k$-NN classifier to achieve high classification speed, (\ref{sec:k-nn}) the pre-computation phase of ClaSP, (\ref{sec:crossval}) the computation of cross-validations for hypothetical split points, (\ref{sec:scoring}) the choice of a scoring metric for a classifier and (\ref{sec:ensembling}) ensembling ClaSPs with temporal constraints (TCs) to increase predictive power. Subsection~\ref{sec:segmentation_alg} describes the application of ClaSP for TSS, (\ref{sec:model-param}) automatically learning proper values for model-parameters $w$ and $C$ and~(\ref{sec:complexity}) studies the computational complexity of our method.

\subsection{$k$-NN Classifier in ClaSP} \label{sec:classifier}

ClaSP uses a $k$-Nearest-Neighbor ($k$-NN) classifier, which has some nice properties that can be exploited to reduce computations when training and evaluating models for every hypothetical split point. Recall that a $k$-NN classifier classifies a given sample by finding the $k$ nearest samples in the training data using a predefined distance function. It then determines the predicted label by aggregating the labels of the $k$-NN training samples. 

\begin{figure}[t]
	\includegraphics[width=1.0\columnwidth]{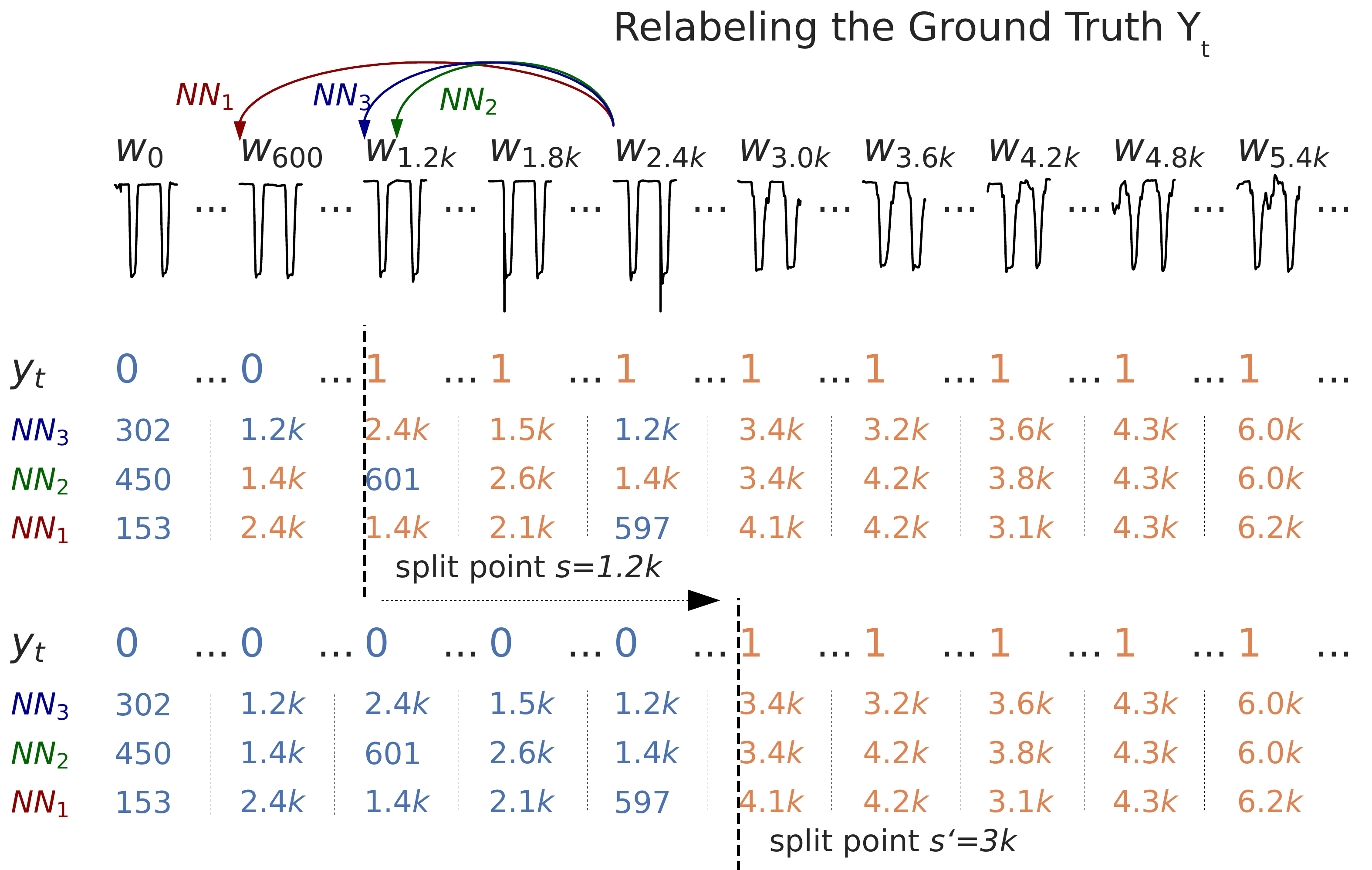}
	\caption{For each window the 3-NNs are computed. When iterating different splits $s$ and $s'$ the ground truth labels $Y_t$ are modified (centre and bottom). The 3-NNs for each window reference new labels, leading to altered predictions for different splits - illustrated in blue and orange for the NNs.
	\label{fig:clasp_relabeling}
	}
\end{figure}

An important property of this procedure is that the pairwise distances between windows and thus the $k$-NNs for any window are \emph{independent of the ground truth labels of the windows}. This means that we need to actually precompute the distance matrix with the $k$-NNs of any window once (Algorithm~\ref{alg:k_nn_profile}). The training and evaluation at a split point $i$ then boils down to looking up the \emph{labels} of the $k$ neighbors in the train split of the cross-validation for every window in the test split, and aggregating them into a majority label. Thus, the expensive distance calculations have to be performed only once, whereas the cheap relabeling is performed on demand for every split point.

Figure~\ref{fig:clasp_relabeling} illustrates this idea for $k=3$. First, $3$-NNs are computed for each window. Note that an NN is represented by the offset of a window, i.e. $302$ for $T_{302,302+w}$. At the left of the split (e.g. 1.2k in the center and 3k in the bottom), we assign class label $0$ (blue), and label $1$ (orange) to the right of the split. For the split at 1.2k (center) there are five windows, two to the left of the split and three to the right that have their $3$-NN windows outside of their own segment, indicated by the orange and blue color. This leads to a classification error, as the classifier predicts the label of the windows in the wrong segment. When moving the split from 1.2k to 3k the ground truth $Y_t$ changes for some windows (but not their distances), indicated by the change in colors. At the same time, the labels of the 3-NN of those windows that were previously mislabeled change in the left and right segment. Thus, for the bottom split, all 3-NN point to windows within the same segments, i.e. class label, resulting in a high classifier accuracy. Actually, once we have pre-computed these $k$-NN offsets for each window, we only need to change the ground truth labels of the NN windows (line~7) to be able to compute a new cross-validation score. Thus, we only have to change one label when moving the split point from $i$ to $i+1$. 

\begin{algorithm}[t]
	\caption{$k$-Nearest-Neighbor Profile}\label{alg:k_nn_profile}
	\begin{algorithmic}[1]
		\Procedure{knn\_profile}{$W$, $k$}
		    \State $\textsc{kNNs} \gets \text{array of shape } \|W\| \times k$
			\State $D \gets \textsc{compute\_distance\_matrix}(W)$ 
            \State $D \gets \textsc{apply\_exclusion\_zone}(D)$ 

			\For{$i \in [1, \dots,  |W|]$}
				\State $A \gets \textsc{argsort}(D[i])$ \Comment{sort row by distance}
				\State $kNNs[i] \gets $ first $k$ offsets from $A$
			\EndFor
			
			\State \Return{$\textsc{kNNs}$}
		\EndProcedure 
	\end{algorithmic}
\end{algorithm}

\subsection{Precompute $k$-NN-Profile} \label{sec:k-nn}

We elaborate on the details of $\textsc{knn\_profile}$ (Algorithm~\ref{alg:clasp} line~5) which pre-computes the $k$-NNs for every window from $T$ using the z-normalized Euclidean distance on the raw values. Algorithm~\ref{alg:k_nn_profile} takes the set $W$ of windows and the $k$ parameter as input, and returns an array of the offsets of the $k$-NNs for each window. First, it computes the pairwise window distance matrix (line~3). It then selects the offsets of those $k$ windows with the lowest distances (lines~5--7), which are returned. We define an exclusion zone around each window of length $w$, so that all windows overlapping with more than $\frac{w}{2}$ points are not considered during the search for NNs (line~4). This effectively sets the distances of windows within $\pm \frac{w}{2}$ of the diagonal of the distance matrix to infinity.
The distance matrix can be computed using the dot-product (line 3). We use a fast implementation, outlined in~\citep{dokmanic2015euclidean, zhu2018matrix}, requiring only $\mathcal{O}(n^2)$ time. Note that also SCRIMP~\citep{zhu2018matrix} implements this idea for computing pairwise z-normalized distances, but only for the overall $1$-NN (and not the $k$-NN as used in ClaSP).

Using symbolic approximations for lower bounding the Euclidean distance like SFA~\citep{SchaferH12} or feature transforms like ROCKET~\citep{dempster2020rocket} is part of our future work.

\subsection{Cross-Validation} \label{sec:crossval}

ClaSP performs a leave-one-out cross-validation for a given offset $i$ using the pre-computed $k$-NN offsets to measure the intra-segment homogeneity of the split, summarized as a classification score. This process is domain-agnostic and introduces only few assumptions, e.g. mutually similar subsequences. Algorithm~\ref{alg:knn_score} illustrates the computation of a single score in ClaSP given the self-supervised ground truth labels $Y_t$ and all $k$-NN offsets. It collects the $k$-NN offsets for each window (line~4), performs a lookup for their current class labels, and finally picks the majority label (line~5). The set of ground truth labels and the set of predicted labels are next passed to a scoring function (line~7). When a split moves, only $Y_t$ changes, potentially resulting in a new score.

\begin{algorithm}[t]
	\caption{Leave-One-Out Cross-Validation Score}\label{alg:knn_score}
	\begin{algorithmic}[1]
		\Procedure{cross\_validate}{$\textsc{kNNs}$,$Y_{t}$}
		
			\State $Y_{pred} \gets $ array of length $\|\textsc{kNNs}\|$, initialized to zero 
			\For{$i \in [1, \dots, \|Y_{pred}\|]$} \Comment{Iterate all windows}
		        \State $\textsc{offsets} \gets \textsc{kNNs}[i]$ \Comment{$k$-NN offsets}
				\State $Y_{pred}[i] \gets $ \textsc{majority\_label}($Y_{t}[\textsc{offsets}]$)
			\EndFor 
			\State \Return{$\textsc{scoring\_function}(Y_{t}, Y_{pred})$}
		\EndProcedure
	\end{algorithmic}
\end{algorithm}

\subsection{Classification Score Selection} \label{sec:scoring}

\begin{figure}[t]
	\includegraphics[width=1.0\columnwidth]{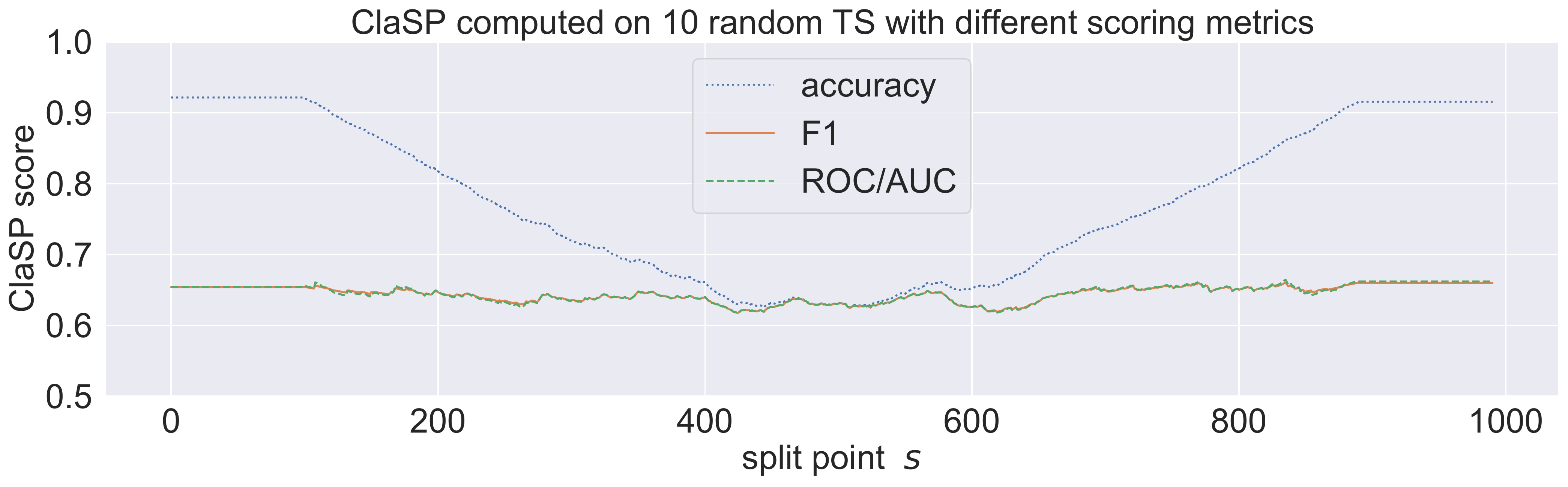}
	\caption{The averaged ClaSP computed from $10$ randomly generated TS, i.e. without homogenous regions. The ClaSP using the optimal score should result in a straight line to avoid any bias, which is the case for F1 and ROC/AUC.\label{fig:ClaSP_random}
	}
\end{figure}

ClaSP determines change points by finding local maxima in the classification score profile. To identify the most disruptive event we need to identify the global maximum (see Figure~\ref{fig:clasp_labeling}d); for solving the problem of TSS, we need to find the $C-1$ highest local maxima, when $C$ is the desired number of segments (see Section~\ref{sec:segmentation_alg}). However, special care has to be taken to make scores of different splits comparable, because for every potential split point in a TS the number of windows in the left and right segments differs. Accordingly, the binary classification problems are often highly class-imbalanced. The degree of class-imbalance changes over the TS, with more severe imbalance towards the ends of the TS and low imbalance at the center. As class imbalance influences the performance of classifiers, a bias may emerge making the comparison of scores of different splits difficult\footnote{Note that FLOSS also has this issue which it solves using score normalization}. 

The impact of this bias depends on the particular evaluation metric being used. To illustrate this, we built a macro-averaged ClaSP from the profiles of $10$ randomly generated TS and plotted the results of different evaluation metrics. These random TS should not contain homogenous segments, thus the optimal evaluation function should result in a constant line. Figure~\ref{fig:ClaSP_random} shows results for accuracy, F1 and ROC/AUC. Accuracy is highly sensitive to class-imbalances and thus leads to a profile similar to a parabola, with high values on the left and right corners. F1 and ROC/AUC both perform much better as shown by the flat line even at the corners of ClaSP. This leads to the conclusion that macro F1 and ROC/AUC are the most appropriate choices for ClaSP.

\subsection{Detecting Reocurring Segments} \label{sec:ensembling}

ClaSP is based on the assumption that the correct hypothetical split point separates two statistically distinguishable parts of a TS. This clearly is the case when a TS has only two segments. However, this assumption can be hurt when three or more segments exists in the data. In this case, ClaSP in its first step, when it determines the first CP, will necessarily treat different segments as one; for instance, in a TS with three segments, the first split will separate one from the other two, which will be treated as if they come from the same class. Unfortunately, having more than one segment within a class increases the intra-class variability, which makes finding the CP more difficult. A further problem comes up if states repeat in the process, leading to reoccurring segments in the data set, like a sequence 'ABA' of two different states 'A' and 'B'. In such cases, ClaSP will often assign different labels to the different instances of the reoccurring segments. This decreases the predictive power of the $k$-NN classifier. 

\begin{algorithm}[t]
	\caption{Classification Score Profile Ensemble}\label{alg:clasp_ensemble}
	\begin{algorithmic}[1]
		\Procedure{calc\_clasp\_ensemble}{$\textsc{T}$, $w$, $n\_iter$}		
			\State $\textsc{ClaSP} \gets \textsc{calc\_clasp}(T, w) $

			\For{$i \in [1, n\_iter] $}
                \State $S, TC \gets $ \textsc{uniform\_sample}($\|T\|$), \textsc{uniform\_sample}($\|T\|$)
			    \State $\textsc{ClaSP}' \gets \textsc{calc\_clasp}(T_{S:S+TC}, w)$
			    \State $\textsc{ClaSP}' \gets \frac{1}{3} \cdot (2 \cdot \textsc{ClaSP}' +
			    \frac{TC}{\|T\|}) $
			    \State $\textsc{ClaSP}_{S:S+TC} \gets \textsc{max}(\textsc{ClaSP}_{S:S+TC}, \textsc{ClaSP}')$
			    
			\EndFor 
			\State \Return{$\textsc{ClaSP}$}
		\EndProcedure
	\end{algorithmic}
\end{algorithm}

A simple idea to tackle this problem is to introduce a temporal constraint (TC) that bounds the maximal gap between a subsequence and its allowed NN in a TS. If the TC is sufficiently small, the NNs are located in the same segment (not in the repeated one) as the subsequence and hence have the same label which leads to a correct classification. FLOSS, for example, bounds its arc curve with a TC that has to be set by a domain expert \citep{gharghabi2017matrix}. The main disadvantage of this idea is that the TC is static. However, TS segments typically vary in subsequence length, such that a pre-defined TC often is either too small or too large and thereby disregards parts of a segment or includes parts of its repetition.

A more sophisticated idea is to use a weighting scheme in the NN determination. Instead of only relying on the z-normalized Euclidean distance as a measure of subsequence similarity, we introduce an error term that considers the proximity of the NNs to the window under consideration. Such a scheme can be understood as a ''soft TC'', as it makes NNs in the local neighborhood of a subsequence more likely to be chosen compared to those further away. ESPRESSO, an extension of FLOSS, uses a weighted arc curve that considers temporal gaps \citep{Deldari2020ESPRESSOEA}. Similar to the TC, however, the major shortcoming of this approach is the fact that the true segment lengths are unknown and neither static nor soft TCs can adapt to their shape.  

To mitigate this problem, we propose to ensemble ClaSPs for randomly selected TCs and aggregate the profiles before taken a decision upon CPs. The main insight underlying this approach is that the quality of a TC for a split correlates with its classification score. A TC too small obstructs the NN determination and a TC too large finds NN in reoccurring segments. Both cases lead to diverse ClaSPs with low scores. Only a well-suited TC that covers large fractions of segments without their repetition produce ClaSPs with distinctive local maxima. In theory, we could consider all possible TCs for a split and choose the one that maximizes the associated classification score. This is, however, computationally expensive and not necessary in practice, as a set of TCs is sufficient to capture parts from a TS that lead to highly predictive splits. Therefore, we randomly draw a predefined amount of TCs, calculate ClaSPs for the TS intervals and aggregate the profiles by maximizing the scores.

Algorithm~\ref{alg:clasp_ensemble} shows the pseudocode for the ensembling. It takes a time series TS $T$, a window size $w$ and the amount of iterations $n\_iter$ as input.  In Subsection~\ref{sec:design_choices}, we show that few iterations are sufficient to detect reoccurring segments and fix $n\_iter$ to a small default value. We start by calculating ClaSP for the entire TS (line~2). This global profile is sufficient to capture the CPs for distinct segments. For each iteration, the procedure randomly samples a start position $S$ and a temporal constraint $TC$ from a uniform distribution and calculates a local ClaSP for the interval $T_{S:S+TC}$ (line~4--5). If the selected interval is larger then $T$, we limit its right end to the last TS data point. The random interval selection allows us to neglect a specific strategy that may favor some segmentations over others. We weigh the local ClaSP with its $TC$ length (line~6) which adds a confidence term to the classification scores that favors larger profiles over smaller ones to prevent overfitting to small TCs. The algorithm updates the global ClaSP in the selected interval by maximizing values with the local ClaSP. This operation stores the best-performing splits in the global ClaSP, which the algorithm eventually reports (line~9). From now on, we refer to the ensembled ClaSP simply as ClaSP.

\begin{figure}[t]
	\includegraphics[width=1.0\columnwidth]{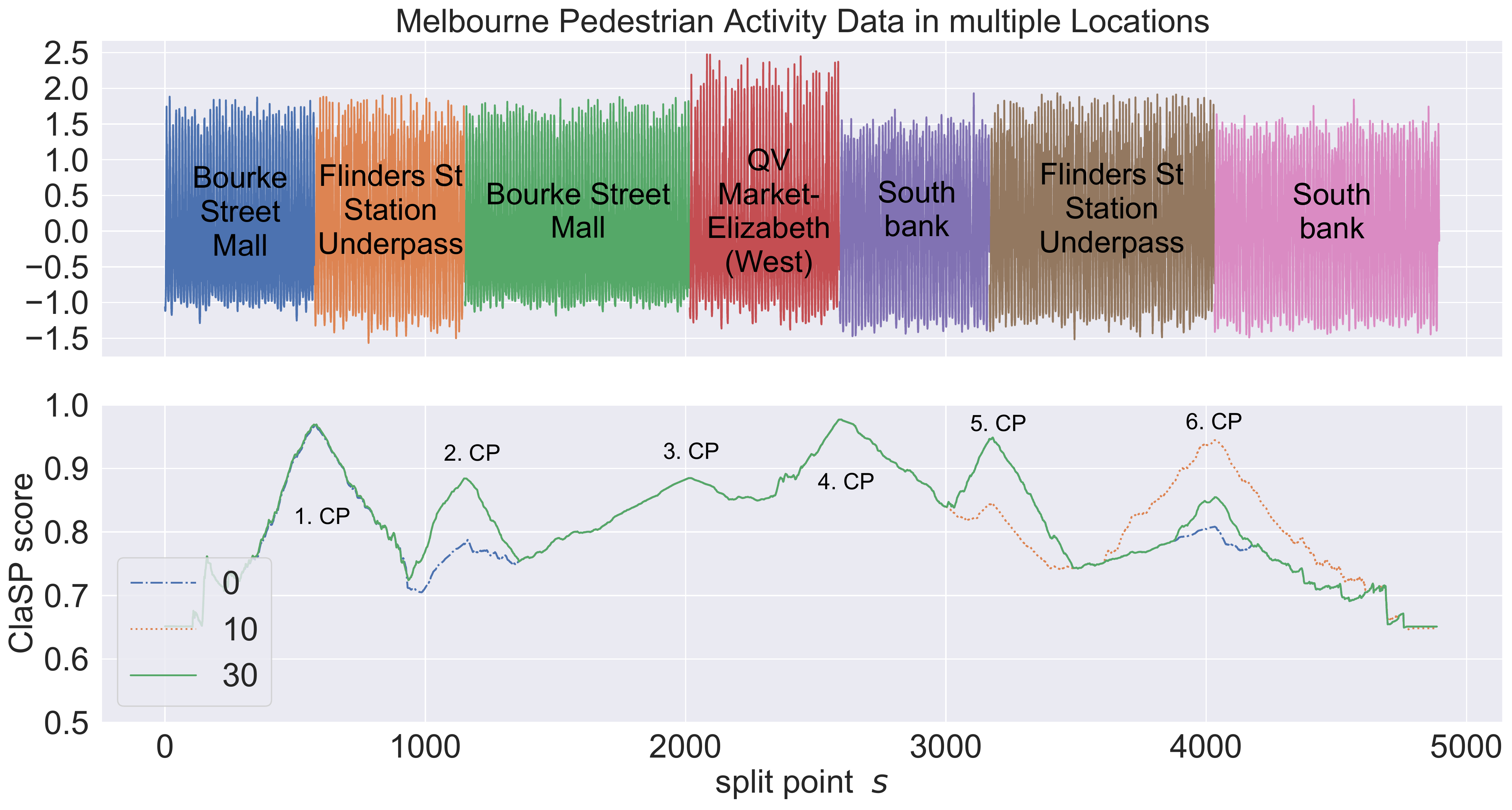}
	\caption{The ensembled ClaSPs with 0, 10 and 30 iterations for a TS with reoccurring sub-segments (ABACDBD). The optimal ClaSP should show substantial peaks at each CP (which is the case for 10 and 30 iterations). \label{fig:ClaSP_ensembling}
	}
\end{figure}

As an example of ensembling see Figure~\ref{fig:ClaSP_ensembling}. The TS (top) illustrates pedestrian activity data from four streets A, B, C, and D in Melbourne as segments~\citep{UCRClassification}. It consists of the sequence of sub-segments A, B, A, C, D, B, and D. Corresponding ClaSPs are computed and depicted with 0, 10 and 30 iterations (see Figure~\ref{fig:ClaSP_ensembling}, bottom). The profiles align for most splits but show distinctive differences at the CPs. While a ClaSP with 0 iterations does not capture all CPs (compare 2nd and 6th transition), 10 and 30 iterations show substantial local maxima for all CPs.

\subsection{Segmentation}\label{sec:segmentation_alg}

Using ClaSP for the detection of a single CP is straightforward: We first compute the profile and then choose its global maximum. Application of ClaSP to cases with more than two segments requires more thoughts. Let us assume one would want to segment a TS $T$ into $C$ segments. A naive idea is to compute ClaSP once and then return the $C-1$ highest validated scores as change points. However, these are typically no local optima, but instead all very close to the highest score (global maximum). Moreover, such approaches depend on the amount of segments $C$ as an input parameter, which may not be known a priori. Another idea is to use a peak finding algorithm and then choose the $C-1$ highest peaks. See~\citep{Yang2008ComparisonOP} for a survey. However, this requires determining parameters of the peak finder, such as minimal gap between peaks or minimal elevation over the local neighborhood. Such information is data-dependent and cannot be easily defaulted.

\begin{algorithm}[t]
	\caption{Segmentation}\label{alg:recsplit}
	\begin{algorithmic}[1]
    	\Procedure{add\_valid\_cpt}{$T$, $w$, $\textsc{pq}$}  \Comment{compute ClaSP, store valid CP}
    	    \State $profile$ $\gets$ \textsc{calc\_clasp\_ensemble($T$, $w$)}
    	    \State ($cpt\_idx, cpt\_val$) $\gets$ (\textsc{argmax}($profile$), \textsc{max}($profile$))
    	    
    		\If{\textsc{validate}($cpt\_val$)} \Comment{Wilcoxon rank-sum test}
        	    \State \textsc{pq.insert}($cpt\_val, (cpt\_idx, T)$)
    	    \EndIf
    	\EndProcedure
    	
		\Procedure{segmentation}{$T$, $w$}
		\State $cpts$ $\gets$ \text{initialize list}
		\State \textsc{pq} $\gets$ \text{initialize max priority queue}
	
		\State \textsc{add\_cpt}($T$, $pq$)

		\While{!\textsc{pq.empty()}} \Comment{process all $C-1$ validated CPs}
			\State ($cpt\_idx$, $S$) $\gets$ \textsc{pq.removeMax()}
			\State $cpts$.\textsc{append}($cpt\_idx$) \Comment{store CP}
			
			\State $T_L \gets S_{begin:cpt\_idx}$ \Comment{calculate CP's left/right segment}
			\State $T_R \gets S_{cpt\_idx+1:end}$
			
			\State \textsc{add\_valid\_cpt}($T_L$, $w$, \textsc{pq}) \Comment{add left/right (valid) CPs}
			\State \textsc{add\_valid\_cpt}($T_R$, $w$, \textsc{pq})
		\EndWhile
		\State \Return{$cpts$}
		\EndProcedure
	\end{algorithmic}
\end{algorithm}

Instead, we propose a novel \emph{parameter-free} strategy based on a recursive splitting of the TS in combination with statistical change point validation. Given $T$ and a window size $w$, the algorithm first computes ClaSP and selects the maximal peak as the first change point if it passes the Wilcoxon rank-sum test (as outlined in Subsection~\ref{sec:cp-candidate-val}). Next, the procedure computes two new ClaSPs, one for the left and one for the right segment of the first split. Within these profiles, it checks and picks the largest peak of both profiles as the second change point. It then computes ClaSP for the two new segments, computes the local maxima, verifies and chooses the highest, etc. This process is recursively repeated until no more valid change points can be extracted. Note that in every iteration, only two ClaSPs have to be computed, as all but one segment remain unchanged compared to the last iteration.

\begin{figure}[t]
	\includegraphics[width=1.0\columnwidth]{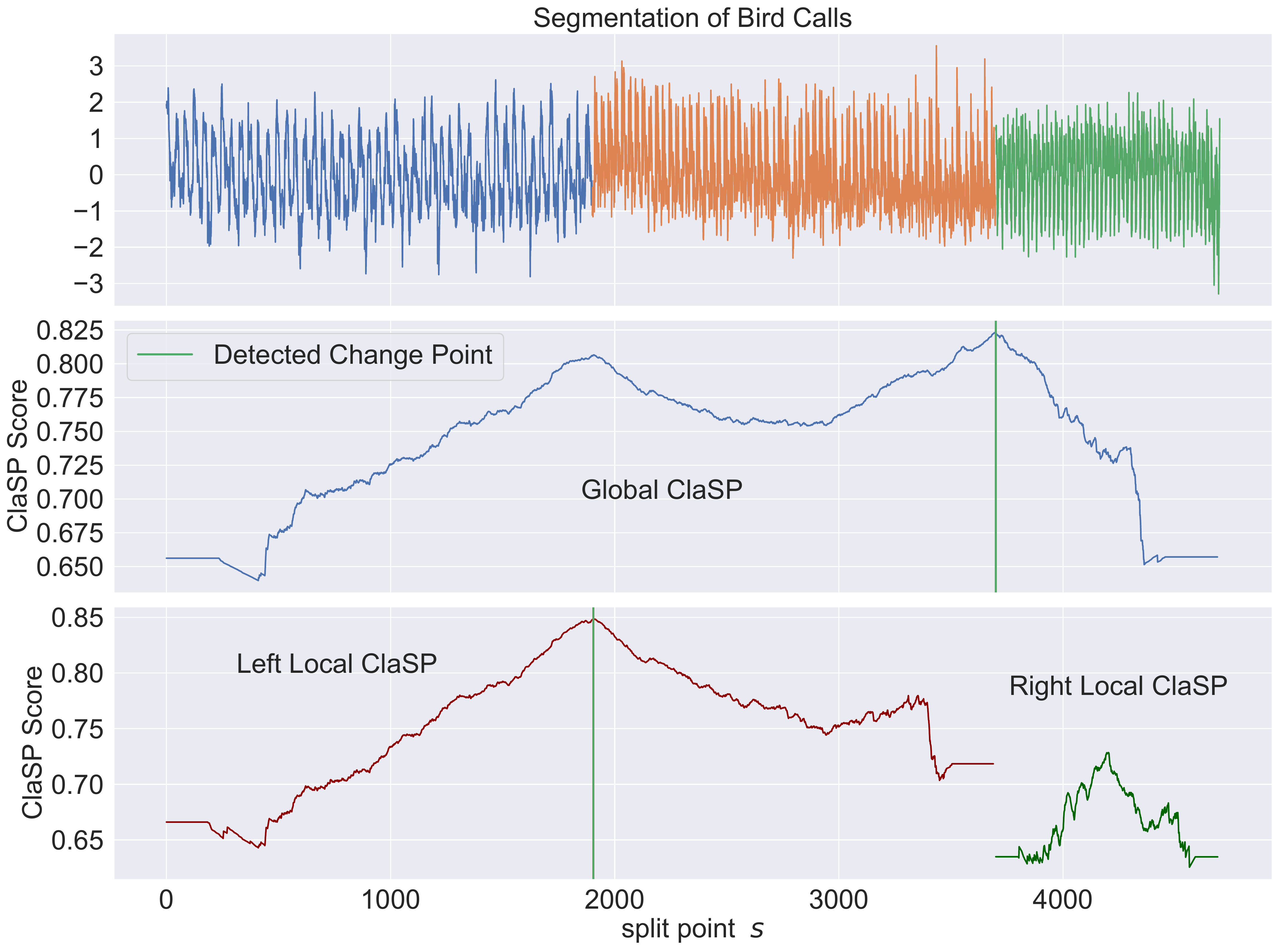}
	\caption{ClaSP is computed, and its global maximum defines the first change point. For each resulting disjoint segment, a (local) ClaSP is computed, and its global maximum is chosen. This process is recursively repeated until the required number of change points (i.e. segments) is reached. \label{fig:ClaSP_recursive_split}
	}
\end{figure}

Pseudocode is shown in Algorithm~\ref{alg:recsplit}. It first calculates the ClaSP over the entire $T$ and stores the index of the peak with the highest score in a priority queue if it passes validation (line~11). Within a loop~(line~13), the offset of the largest local maximum $cpt\_idx$ is extracted and the two left and right segments of the split $cpt\_idx$ are derived (lines~15--16), for which the corresponding ClaSPs are computed, their potential CPs checked and added to the priority queue if confirmed (lines~17--18). The loop ends when no further CP can be approved.

An example is shown in Figure~\ref{fig:ClaSP_recursive_split}. The TS (top) shows three different bird calls from the Great Barbet, Megalaima virens as segments (taken from~\citep{gharghabi2017matrix}). The TS consists of the 7th Mel Frequency Cepstral Coefficient of the recordings with which the individuals can be identified. The first profile (Figure~\ref{fig:ClaSP_recursive_split}, center) already contains clear peaks for both CPs, but the second peak becomes more precise through the second iteration. The main advantages of this iterative process is that it is parameter-free and does not require to set an exclusion zone around peaks.

\subsection{Learning the Model-Parameters} \label{sec:model-param}

ClaSP has two domain-dependent model-parameters, the window size $w$ and the amount of segments $C$. Learning these parameters automatically makes the problem of TSS much harder, as no expert knowledge can be exploited. However, it enables the automatic application of TS technology in challenging scenarios with little prior knowledge about the observed process at hand. In the following subsections, we introduce two novel domain-agnostic techniques to select proper window sizes from TS and to detect the number of CPs in data sets. We use these methods to automatically learn the model-parameters in ClaSP. 

\subsubsection{Selecting the Window Size} \label{sec:window-size-selec}

In the literature, it is common to treat the window size as a hyper-parameter to be set manually by a domain expert~\citep{gharghabi2017matrix}. This implicitly allows for the usage of background knowledge, but is costly and slow, as such experts first have to be identified. ClaSP learns the window size $w$ as a model-parameter from the data. This parameter has data-dependent effects on its performance. When chosen too small, all windows tend to appear similar; when chosen too large, windows have a higher chance to overlap adjacent segments, blurring their discriminative power. It is thus non-trivial to pick a suitable window size for a given TS. The ideal window size should correlate with the inherent statistical properties of the given data set.

\begin{algorithm}[t]
	\caption{Summary Statistics Subsequence}\label{alg:suss}
	\begin{algorithmic}[1]
		\Procedure{stats\_diff}{$T$, $w$, $stats_T$}		
			\State $stats_w \gets (roll\_mean(T, w), roll\_std(T, w), roll\_range(T,w))$
		    \State $stats_{diff} \gets \frac{1}{\sqrt{w}} \cdot $ \textsc{euclidean\_distance}($stats_T, stats_w$)
		    \State \Return{$mean(stats_{diff})$}
		\EndProcedure
		
	    \Procedure{suss\_score}{$T$, $w$, $stats_T$}		
			\State $s_{min}, s_{max} \gets $ \textsc{stats\_diff($T, \|T\|, stats_T$)}, \textsc{stats\_diff($T, 1, stats_T$)}
			\State $score \gets $ min-max scale \textsc{stats\_diff($T, w, stats_T$)} to $[s_{min},\dots,s_{max}]$
			\State \Return{$1-score$}
		\EndProcedure
		
		\Procedure{calc\_suss}{$T$, $t$}		
			\State $T \gets $ min-max scale $T$ to range $[0,\dots,1]$
			\State $stats_T  \gets $ $(mean(T), std(T), 1)$
			
			\State $lbound, ubound \gets $ \textsc{exponential\_search($T, stats_T,$\textsc{suss\_score}$,t$)}
			\State $w \gets $ \textsc{binary\_search}($T, stats_T,$\textsc{suss\_score}$,lbound, ubound, t$)
			\State \Return{$w$}
		\EndProcedure
	\end{algorithmic}
\end{algorithm}

We elaborate ways to infer the window size automatically from the data. A prominent approach is periodicity detection that tries to detect a TS' most dominant periods \citep{Elfeky2005PeriodicityDI}. One technique to extract periods is to select the most dominant coefficients of a Discrete Fourier Transform and to transform the corresponding frequency to a period size. Another method to extract periods is to calculate the Auto Correlation Function (ACF) of a TS, which reports the correlation of the data set with a delayed copy of itself. The lag with the highest correlation is the most dominant period size~\citep{Vlachos2005OnPD, Wen2021RobustPeriodRT}. One major drawback of periodicity detection is that it can only work if a data set actually has detectable periods, which must not necessarily be the case.

\begin{figure}[t]
	\includegraphics[width=1.0\columnwidth]{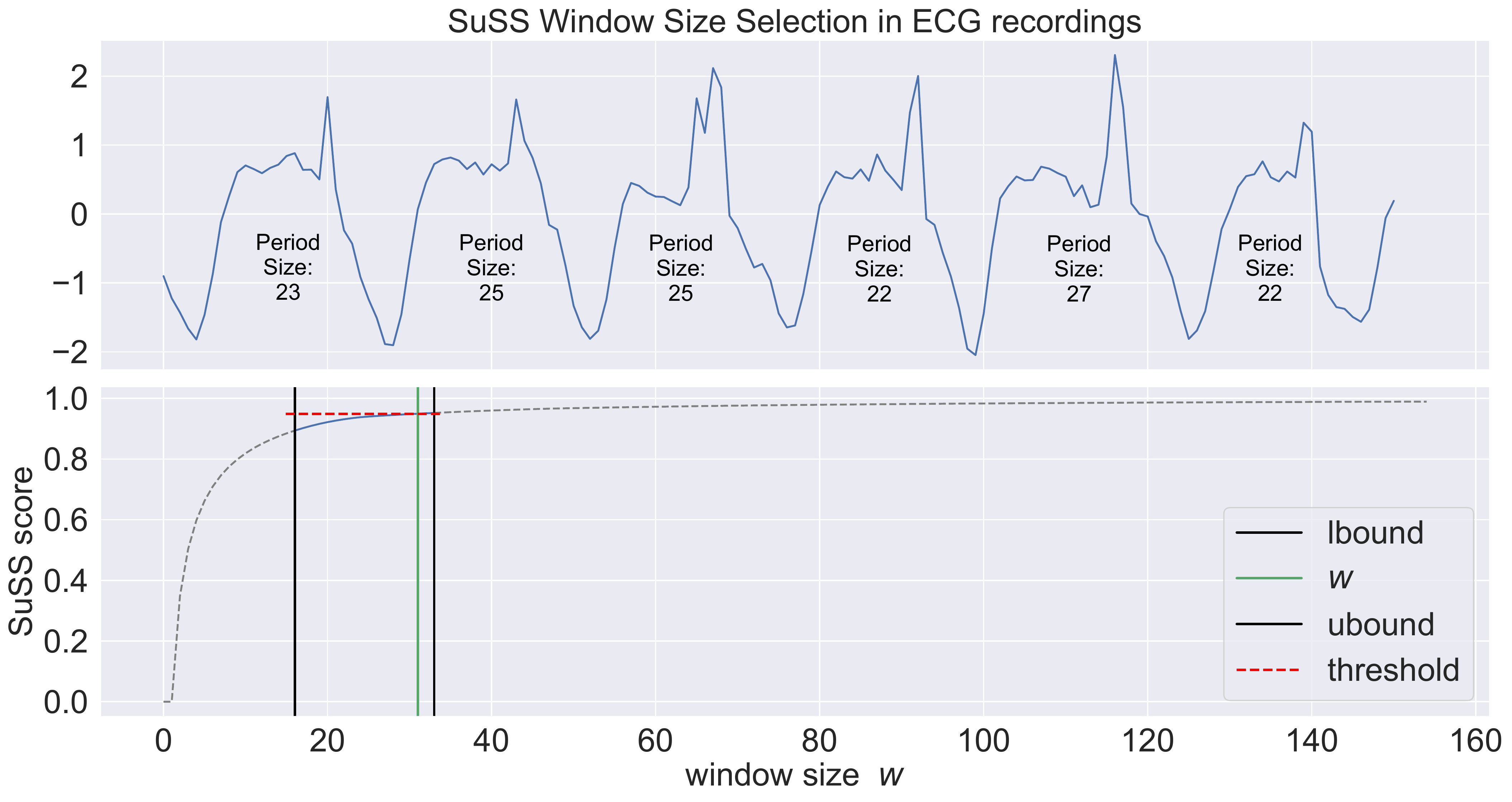}
	\caption{SuSS locates a narrow range between $lbound$ and $ubound$ (using exponential search) in which the requested window size $w$ is found (using binary search). The SuSS scores  monotonically increase and converge for windows containing more than 40 data points. \label{fig:SuSS_curve}
	}
\end{figure}

We propose \textit{Summary Statistics Subsequence (SuSS)}, a novel time series window size selection algorithm that compares local subsequences with the global TS using descriptive statistics. The central hypothesis of SuSS is that the summary statistics of subsequences with optimal size should be close to the ones of the entire data set. Therefore, such windows are representative for the time series and should be suitable for classification in ClaSP. SuSS takes a TS $T$ as input and calculates its mean, standard deviation and range as a summary statistics vector $stats_T$ of size 3 (Algorithm~\ref{alg:suss}, line~13). We then use these global statistics to calculate their distance to the rolling statistical sizes of windows with increasing length. For a candidate window size $w \in [1, \|T\|]$, the rolling summary statistics $stats_w$ are calculated as a matrix such that the $i$-th row contains the statistics for $T_{i,i+w}$ (line~2). SuSS calculates the Euclidean distances between $stats_T$ and all rows in $stats_w$ and weighs them with the inverse of the root of the window size (line~3). We normalize the distances in order to correct a bias for larger windows that are inherently closer to the TS. This step creates length-invariant distances that are comparable across window sizes. The mean distance (line~4) is scaled and represents the final SuSS score for $w$ (line~7--9). 

Naturally, the scores monotonically increase with greater $w$ as the statistics of the windows and the TS align. Therefore, we conduct an exponential and binary search to efficiently find the smallest window size $w$ with a score larger than a pre-defined threshold $t \in [0,\dots,1]$. As we search for the smallest window size for which the windows' summary statistics closely match the ones of the entire TS, we fix the threshold $t$ to a domain-agnostic default value in Subsection~\ref{sec:design_choices}. The exponential search spans a small interval $lbound \leq w \leq ubound$ in which the binary search locates $w$ which SuSS eventually reports (line~14-16). The combination of both search procedures ensures a fast window size selection with few comparisons. 

Figure~\ref{fig:SuSS_curve} shows an example. The TS (top) is an excerpt of electrical activity of normal human heartbeats (taken from~\citep{UCRClassification}). For illustration, all the SuSS scores for window size 1 to 160 are computed (Figure~\ref{fig:SuSS_curve}, bottom). The scores quickly converge as the optimal window size is independent of the TS length. For a threshold of 95\%, SuSS firstly bounds the search space between 16 and 32 and then finds the window size 31 (score 95,1\%) which accurately captures a single heartbeat (20-30 data points) that closely matches the summary statistics of the entire data set. In the experimental evaluation, we compare SuSS with competitor methods. 

\subsubsection{Detecting the Number of Change Points} \label{sec:cp-candidate-val}

Local maxima in ClaSP reflect potential change points. By visual inspection, a domain expert may be able to identify local maxima that correspond to CPs in easy cases. However, it is unclear how substantial a deflection has to be to qualify as a state change in the underlying process. In some problem settings, the amount of segments $C$ in a TS may be known a priori because of domain knowledge. A segmentation procedure (as outlined in Section~\ref{sec:segmentation_alg}) can then simply extract the first $C-1$ most dominant CPs and terminate. However, in general the hyper-parameter $C$ is not known but should be determined by the method itself.

We tested multiple simple baselines to learn the number of change points from the data. A simple idea to tackle this problem is to set a classification score threshold and disregard CP candidates with low scores. We expect CPs to correlate with high predictive performance in ClaSP. Thus, if we set a high threshold, e.g. 75\%, for the ROC-AUC score, we should have sufficient confidence to only report CPs that equate to real state changes in the physical process. The main shortcoming of this approach is, however, that the performance of the $k$-NN classifier in ClaSP is data-dependent. Hence, a static threshold can be too conservative for time series with low separability or too liberal for ones with high variability. 

A more adaptive approach is to set a threshold that bounds the gini gain of a CP candidate~\citep{Breiman2004TechnicalNS}. The gini impurity, in our context, reports the probability of misclassifying a randomly chosen subsequence in a hypothetical split. The gini gain subtracts the weighted segment impurities (left and right) from the splits' impurity (combined). The computed value is high for splits that create pure segments and medium to low for more diverse ones. If we only consider CPs that fulfil a pre-defined gini gain, we can enforce an increase in homogeneity for new segments. Similar to the static classification score thresholding, however, a key disadvantage of this idea is that a pre-defined gini gain may be more or less suitable for some TS.

In contrast to previous works as described in~\citep{aminikhanghahi2017survey,truong2020selective} and simple baselines, ClaSP uses hypothesis testing to determine statistically significant change points. We use the non-parametric two-sided Wilcoxon rank-sum test to validate potential CPs. It checks that for an investigated split, the predicted labels from the left segment are more likely to be smaller than the ones from the right segment for a pre-defined confidence. Recall that, by definition, a true predicted CP creates a pure label vector that mostly contains zeros in the left segment and ones in the right one. To detect significant CPs, we simply fix a robust maximal approved p-value for ClaSP that needs to be passed (see Subsection~\ref{sec:design_choices}). This approach has multiple benefits in comparison to the two aforementioned procedures. Firstly, the maximal approved p-value is independent of the used classification score and easily interpretable by humans. Secondly, the test statistic directly checks the probability of a split corresponding to a CP without using a proxy like the gini gain that introduces further assumptions. Thirdly, it is non-parametric and thus applicable to label configurations from any distribution. 

\begin{figure}[t]
	\includegraphics[width=1.0\columnwidth]{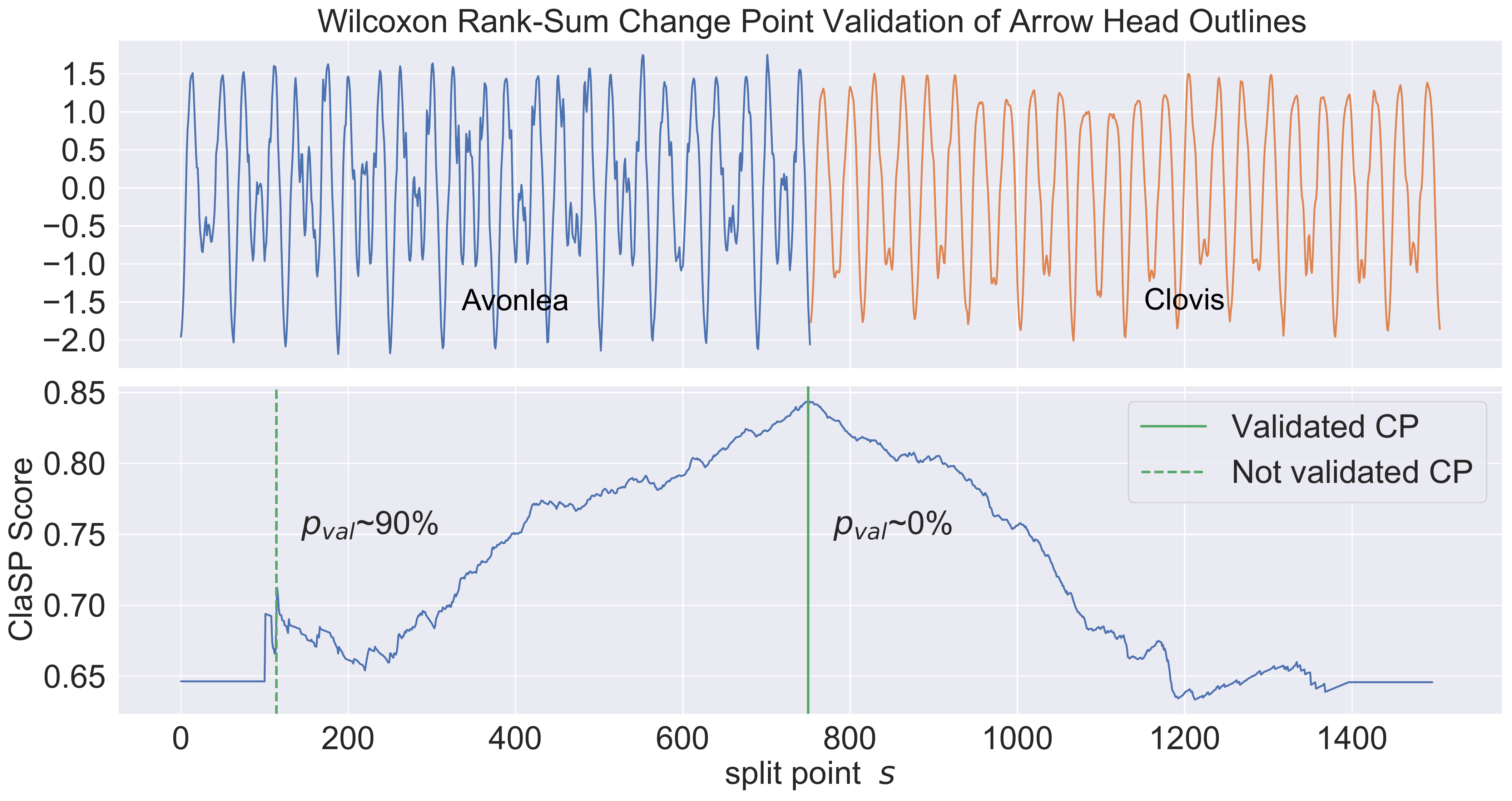}
	\caption{The global maximum of the computed ClaSP passes the Wilcoxon rank-sum test with small a p-value close to 0\% and a threshold of 5\%. The second-highest local maxima does not pass the validation with a high p-value of 90\%. \label{fig:ClaSP_cp_validation}
	}
\end{figure}

Figure~\ref{fig:ClaSP_cp_validation} shows an example of change point candidate validation using the Wilcoxon rank-sum test. The TS (top) illustrates instances of different arrowhead shapes as segments~\citep{UCRClassification}. The corresponding ClaSP (Figure~\ref{fig:ClaSP_cp_validation}, bottom) shows a clear global maximum (split 750, score 84.4\%) that divides the two different arrowhead types and one smaller local maximum (split 114, score 69\%) in the beginning of the TS. Although both associated scores are medium to high, the probability that the splits are not CPs is close to zero for the global maximum and very high for the local maximum, which reflects the ground truth. In the experimental evaluation, we compare the performance of the different approaches.

\subsection{Computational Complexity}\label{sec:complexity}

In this Subsection, we analyse the computational worst-case complexity of the individual steps of ClaSP and eventually derive its overall complexity. 

In order to compute ClaSP, we first need to determine its window size using \textsc{calc\_suss} (Algorithm~\ref{alg:suss}). For a TS of length $n$ and a candidate window size, this computation mainly depends on calculating the rolling window statistics and their distance to the global TS statistics, which can be computed in $\mathcal{O}(n)$ (line 2--3). Searching the candidate space is $\mathcal{O}(n \cdot \log w)$ for binary and exponential search (line 14--15). As the window size is constraint by $n$, SuSS has a worst-case complexity of $\mathcal{O}(n \cdot \log n)$. In most applications, however, $w$ is a small constant which leads to a runtime complexity of $\mathcal{O}(n)$.

The computational complexity of \textsc{calc\_clasp} (Algorithm~\ref{alg:clasp}) using a $k$-NN classifier is dominated by the cost of \textsc{knn\_profile} (line~5) and $\mathcal{O}(n)$ calls to \textsc{cross\_validation} (line~6--9). The \textsc{knn\_profile} calculates the distance matrix (line~4), which can be computed in $\mathcal{O}(n^2)$ \citep{dokmanic2015euclidean, zhu2018matrix}. Retrieving the $k$ smallest window offsets (line~8) can be solved in $\mathcal{O}(k \cdot n)$ using $k$ sequential searches or $\mathcal{O}(n + k \log n)$, when using a min-heap. As $k$ is a (small) constant, line~5--8 are performed in $\mathcal{O}(n^2)$. \textsc{cross\_validation} can be performed in $\mathcal{O}(n)$ (line~3--6) for a single split index; as there can be at most $n$ splits, complexity of this step also is $\mathcal{O}(n^2)$. Hence, the total runtime complexity of computing a single ClaSP is in $\mathcal{O}(n^2)$. 

The ensembling of multiple ClaSPs (Algorithm~\ref{alg:clasp_ensemble}) computes one ClaSP per iteration, leading to $\mathcal{O}(n\_iter \cdot n^2)$. With $n\_iter$ being a (small) constant, also the total computational complexity of computing a split point with ClaSP is in $\mathcal{O}(n^2)$. Segmentation into $C$ different segments (Algorithm \ref{alg:recsplit}) requires $C$ such computations (line~17--18), regardless of $C$ being known a priori or being determined during the run of the algorithm. TSS with ClaSP for $C$ segments thus requires altogether $\mathcal{O}(C \cdot n^2)$. 

Note, that the main contributing factors to the runtime are the $k$-NN computations and cross-validations, repeated $n\_iter \cdot C$ times. This limits the scalability of ClaSP and prevents its direct application to an online setting. We consider scalable streaming adaptions of ClaSP as part of our future work.

\section{Experimental Evaluation}\label{sec:experiments}

In this section, we experimentally study the accuracy and runtime of ClaSP on a large set of TS and compare these values to six competitors. We shall first describe the time series and evaluation metrics used for the evaluations (see Subsection~\ref{sec:setup}). We then describe a series of experiments to study the influence of different design choices in Subsection~\ref{sec:design_choices}. We compare ClaSP’s performance and runtime with six state-of-the-art competitors in Subsections~\ref{sec:benchmark_segmentation} and~\ref{sec:runtime}, and show results for ClaSP with complex classifiers (see Subsection~\ref{sec:complex-classifiers}). Finally, we shall discuss three particularly challenging real-life data sets to show features and some limitations of ClaSP in Subsection~\ref{sec:use_cases}. To ensure reproducible results and to foster follow-up works, we provide ClaSP's source code, Jupyter-Notebooks, all TS used in the benchmarks, visualizations, and the raw measurement sheets on our website~\citep{ClaSPWebpage}.

\subsection{Benchmark Setup}
\label{sec:setup}

\paragraph{Data Sets}
Overall we use $107$ benchmark data sets to assess the performance of ClaSP and to compare it to rivaling methods, which, to the best of our knowledge, is the largest collection of data sets (partly human annotated, partly semi-synthetic, partly fully synthetic) considered so far for change point detection. We release the entire Time Series Segmentation Benchmark (TSSB) set on GitHub~\citep{TSSBWebpage} to foster the comparability for future TSS research.

\textbf{UTSA Benchmark Datasets:} $32$ data sets stem from the public UCR Time Series Semantic Segmentation Archive (UTSA)~\citep{gharghabi2017matrix}; these are TS from the literature that capture biological, mechanical or synthetic processes with 2 to 3 unique segments. The TS group into 16 use cases, from which $8$ are data sets with real changes, $7$ include semi-synthetic TS with artificial changes (e.g. concatenated or flipped recordings) and one is fully-synthetic (different sine waves). Some TS were further pre-processed by the authors (e.g. applying a band-pass filter, resampling or clipping). Change points as well as period sizes were annotated by human experts as described in~\citep{gharghabi2017matrix}.

\textbf{TSSB Benchmark Datasets:} The remaining $75$ TS are semi-synthetically created using real-world TS from the UCR archive~\citep{UCRClassification}. From the entire archive set, we first removed all DS with missing values or too few (or too many) TS per class label. We finally selected a subset of $75$ data sets that have in common that whole TS per class are similar to each other, such that their entirety shows some obvious periodicity (compare our assumptions in Section~\ref{sec:background}). For each of these, we then created a segmentation problem as follows: We first manually choose a subset of classes (e.g. 3 out of 5) and an order of their occurrence (e.g. ABC). We then group all TS by class label (in no particular order) and concatenate all TS accordingly, to create segments with repeating temporal patterns. A given segment in the resulting TS thus contains multiple instances of the distinct label characteristics. The location at which different segments (classes) were concatenated are marked as change points. We investigated if the concatenation points introduce substantial gaps but found no significant difference compared to other time points. While most of the resulting TS, we created, contain 2-9 unique segments, we also include $6$ TS with only 1 segment (no CP) and $10$ TS with reoccurring sub-segments. For this, we cut segments into 2 or 3 parts and manually position them throughout the TS (e.g. ABAC). We resample the resulting TS using mean values to control their resolution and we annotate the data set with window sizes that are hand-selected from the values $\{10,20,50,100\}$ to capture temporal patterns. We create diverse problem settings with a controlled  amount of segments and varying resample rates to create a wide variety of TS, e.g. short (or long) TS with few (or many) segments. This process requires human supervision to check that the TS resolution maintains interpretable temporal patterns and that the segment ordering does not introduce large gaps. We only keep high quality TS where these requirements are met. By definition, however, dataset creation has a selection bias. Therefore, we release all of our design choices and a Jupyter Notebook to automatically reproduce the TSSB and the code to replicate own variants on GitHub~\citep{TSSBWebpage} for transparency.

Out of all $107$ data sets, the smallest (largest) ones have 240 (40k) data points (median 5196). $6$ TS have $1$ segment (no CP), $46$ TS have $2$ segments (1 CP), $20$ data sets have $3$ segments, $15$ data sets have $4$ segments, $11$ data sets have $5$ segments, $2$ data sets have $6$ segments, $6$ data sets have $6$ segments, and $1$ data set has $9$ segments. The smallest to largest annotated window sizes span from 10 to 400 data points (median 12). We use these pre-defined values for all methods that set the window size as a hyper-parameter. In Subsection~\ref{sec:design_choices}, we will perform experiments with these values being pre-defined as well as determined by ClaSP itself to compare the impact of different window size selection strategies.

\paragraph{Competitors}
We compare ClaSP to six state-of-the-art competitors. Two of the competitors are the NN-based approaches FLOSS~\citep{gharghabi2017matrix} and ESPRESSO~\citep{Deldari2020ESPRESSOEA}. FLOSS does not specify how to learn the number of CPs automatically. For experiments without pre-defined $C$, we learned an optimal threshold for its arc curve (0.45) and only report extracted CPs with smaller or equal scores. ESPRESSO sets a chain length (3) and a maximum amount of CPs (8) as hyper-parameters. The next three methods are best-performing methods according to the comparison published in~\citep{van2020evaluation}, namely Binary Segmentation (BinSeg)~\citep{scott1974cluster} Bayesian Online Changepoint Detection (BOCD)~\citep{adams2007bayesian} and Pruned Exact Linear Time (PELT)~\citep{Killick2012OptimalDO}. BinSeg and PELT set a cost function and a CP penalty as hyper-parameters. As a cost function, we tested autoregressive, cosine, Gaussian, kernel, L1, L2, and Mahalanobis and as CP penalties we evaluated 10, 20, and $\log T$ (data not shown). We choose autoregressive and 10 as default hyper-parameters for both algorithms. As a sixth method, we include the results for a simple window-based baseline (Window) \citep{truong2020selective} with Mahalanobis cost function and a CP penalty of 20.

\paragraph{Evaluation Metrics}
The literature contains multiple metrics to assess segmentation procedures. They can be divided into classification- and clustering-based approaches. The former emphasizes the importance of detecting the actual timestamps at which the underlying process changes and report F1 values for predicted versus annotated CPs. In contrast, the latter focuses on dividing a TS into homogeneous segments and report a measure for the overlaps of predicted versus annotated segments. We entertain both views and will report both the F1 score as well as the Covering score~\citep{van2020evaluation}. These scores are defined as follows. 

\textbf{F1 score for TSS:} Assume a TS $T$ of length $n$ and sets of ground truth CPs $cpts_{T}$ and predicted CPs $cpts_{pred}$, with each location in $[1,\dots,n]$. The F1 score reports the harmonic mean between precision and recall (see Subsection~\ref{sec:eval_metrics}), where precision reflects the fraction of correctly identified CPs over the number of predicted CPs while recall computes the number of correctly identified CPs over the number of ground truth CPs. It is a common practice to consider predicted CPs as TPs if they are in close proximity to ground truth CPs. Such slack is necessary as the TS size is substantially larger than its amount of CPs. We choose 1\% of the TS size as a margin of error and allow exactly one correctly predicted CP to count as a TP.

\textbf{Covering score for TSS:} Let the interval of successive CPs $[t_{i_k},\dots,t_{i_{k+1}}]$ denote a segment in $T$ and let $segs_{pred}$ as well as $segs_{T}$ be the sets of predicted or ground truth segmentations, respectively. For notational convenience, we always consider $t_{i_0} = 0$ as the first and $t_{i_{C}} = n+1$ as the last CP to include the first (last) segment. The Covering score reports the best-scoring weighted overlap between a ground truth and a predicted segmentation (using the Jaccard index) as a normed value in the interval $[0,\dots,1]$ with higher being better (equation~\ref{eqn:covering}). 
\begin{align}
    Covering = \frac{1}{\|T\|} \sum_{s \in segs_{T}} \|s\| \cdot \max_{s' \in segs_{pred}} \frac{\| s \cap s' \|}{\| s \cup s' \|}
    \label{eqn:covering}
\end{align}

Both metrics allow us to compare sets of $cpts_{T}$ and $cpts_{pred}$ with different lengths, including empty sets. Its typically harder to obtain high F1 scores compared to Covering scores. This is due to the fact that the F1 score considers a margin of error and harshly penalizes CPs that are out-of-range. In the evaluation, we choose to focus on the Covering score and only discuss the substantial differences compared to the F1 (if there are any). To aggregate the results of all methods on the $107$ data sets into a single ranking, we first compute the rank of the score of each method per data set, i.e., the best method is assigned rank 1, the 2nd best to rank 2 etc. We then average ranks of a method on all data sets to obtain its overall rank. We use critical difference diagrams (as introduced in~\citep{demvsar2006statistical}) to compare ranks between approaches. The best approaches scoring the lowest (average) ranks are shown to the right of the diagram; see, for instance, Figure~\ref{fig:cd_unsupervised_seg}. Groups of approaches that are not significantly different in their ranks are connected by a bar, based on a Nemenyi two tailed significance test with $\alpha=0.05$.

\subsection{Ablation Study on Design Choices}
\label{sec:design_choices}

There are five main design choices for ClaSP: (a) the number of $k$ neighbors used for classification, (b) the classification score and its treatment of class imbalance, (c) the number of iterations $n\_iter$ in the ensemble, (d) the selection of the window size $w$, and (e) the change point candidate validation test used to learn the number of segments $C$. We describe the results of extensive ablation studies to fix default values (methods) for options (a--e). Note that ClaSP has no default values for the window size or the number of segments, but rather learns them as model-parameters automatically from the data.

\begin{figure}[t]
	\begin{minipage}{6cm}
        \includegraphics[width=1.0\columnwidth]{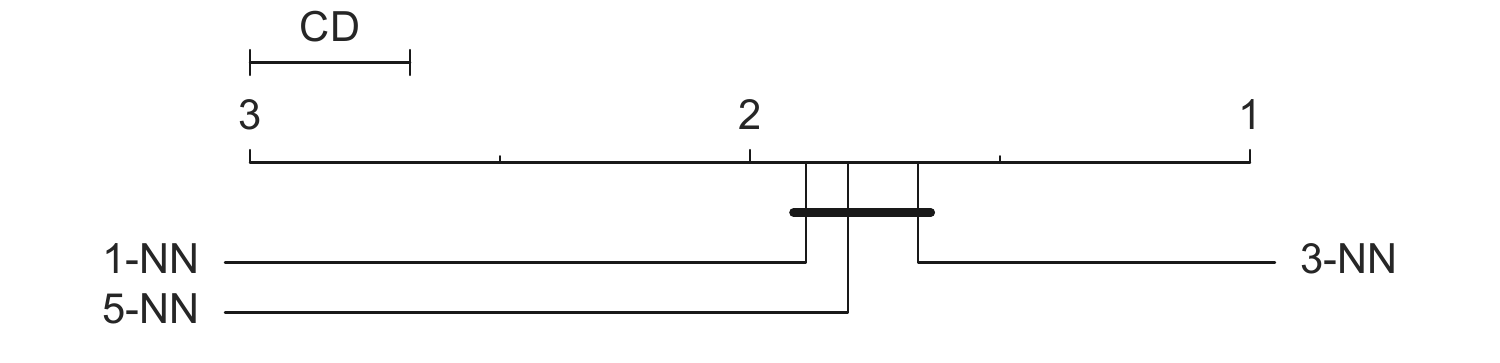}
	\end{minipage}
	\begin{minipage}{6cm}
        \includegraphics[width=1.0\columnwidth]{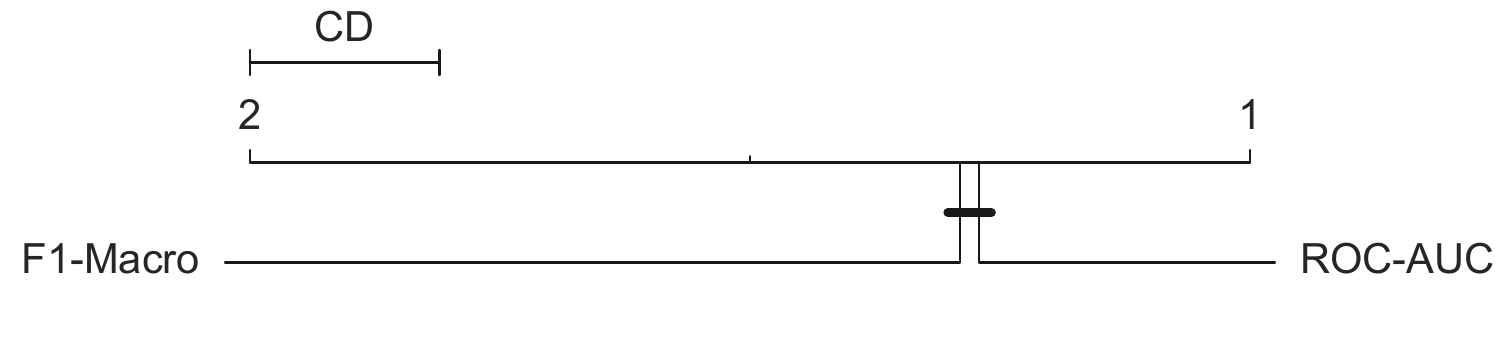}
	\end{minipage}
	\begin{minipage}{6cm}
        \includegraphics[width=1.0\columnwidth]{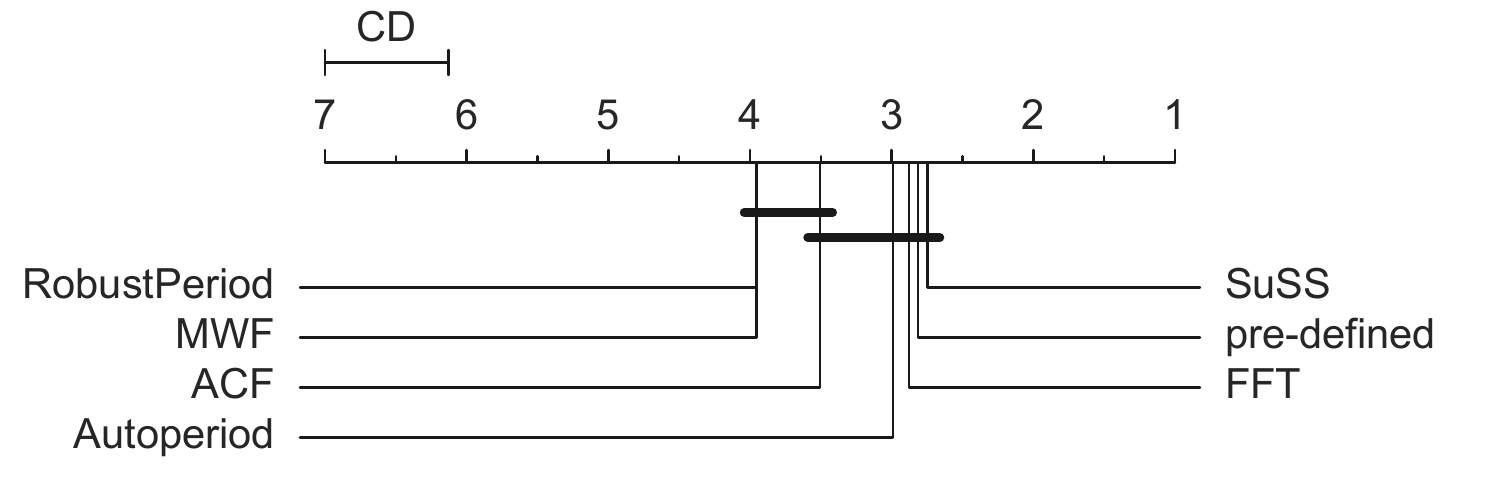}
	\end{minipage}
	\begin{minipage}{6cm}
        \includegraphics[width=1.0\columnwidth]{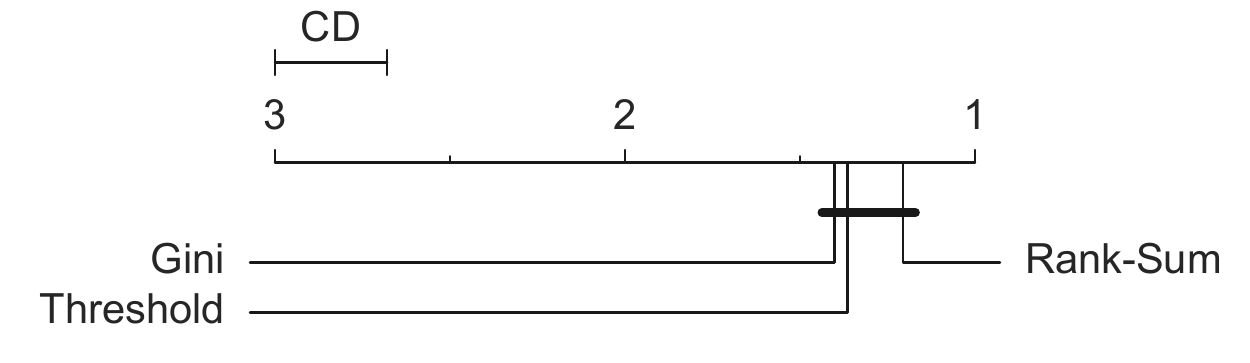}
	\end{minipage}
	\caption{Ablation Study of ClaSP: Influence of the $k$ neighbours parameter (top left), the classification score (top right), window size selection (bottom left), and change point candidate validation (bottom right) on the average Covering rank for the $107$ benchmark data sets.\label{fig:cd_design_choices}
	}
\end{figure}

\paragraph{$k$-NN Classifier and Classification Score}
We tested $k \in \{1,3,5\}$ neighbours for (a) and the F1 score (macro) as well as ROC/AUC score for (b). Figure~\ref{fig:cd_design_choices} (top) shows the critical difference diagram over all configurations computed on the entire benchmark set. Ranks differ only insignificantly for both metrics. We therefore choose $k=3$ and ROC/AUC because this configuration achieves the lowest average Covering/F1 ranks and the highest average performance for both evaluation metrics. A larger $k$ is also less affected by outliers / extreme values.

\paragraph{Number of Iterations in the Ensembling}
We evaluated the number of iterations $n\_iter \in [0,\dots,100]$ (steps of 10) in the ensemble for (c). Overall, we found no significant differences between the average Covering (and F1) ranks of the different values (data not shown). Using no iterations performs worst (average rank 2.36) and as of 30 iterations the average ranks converge between 1.2 and 1.4. Additionally, we tested TS with (and without) reoccurring segments in isolation. For TS with repeating segments, 30-100 iterations perform on average 29.3\% better compared to no iterations. For distinct segments, all variants are on par. Accordingly, we set $n\_iter=30$  as the default value to be able to detect reoccurring segments while limiting the increase in runtime.

\paragraph{Window Size Selection}
Regarding the window size (d), we performed experiments to find a suitable threshold $t$ for SuSS and compared the best-scoring version to window size selection techniques from the literature and to the annotations in the benchmark. We tested $t \in [0.85,\dots,0.95]$ with step size 0.01 (data not shown) and found all thresholds to rank insignificantly different, except 0.94 and 0.95 that score worse. For all thresholds, the mean (std deviation) Covering performances are within 81.4\% and 84.7\% (18.1\% and 21.7\%). We set $t=0.89$ as default because it has the lowest average Covering (and F1) rank. We then compared SuSS with this value to the highest auto-correlation (ACF), Autoperiod, the most dominant Fourier frequency (FFT), pre-defined annotations, the Multi-Window-Finder (MWF) \citep{ImaniMultiWindowFinderDA}, and RobustPeriod for window size selection in ClaSP. For all procedures, we tested the computed window size $w$ as well as $\frac{w}{2}$. Figure~\ref{fig:cd_design_choices} (bottom left) shows the results. SuSS shows the lowest average Covering (and F1) rank, followed by pre-defined annotations, FFT, Autoperiod and ACF. MWF and RobustPeriod rank significantly worse. SuSS and pre-defined annotations are on par considering their average rank (2.7 vs 2.8) and Covering performance (83.8\% vs 83.3\%) as well as std deviation (19.4\% vs 21.1\%). FFT shows similar summary statistics, but ranks worse. We find these results rather interesting, as they indicate that SuSS might make costly pre-defined annotations unnecessary. Therefore, we choose SuSS for window size selection.

\paragraph{Change Point Validation}
Considering the change point candidate validation (e), we determined a default min. score for the ROC/AUC threshold (i) in range $[0.5,\dots,0.95]$ (steps of 0.05), a min. gini gain (ii) in range $[0.005,\dots,0.05]$ (with step size 0.005), and a max. approved p-value for the Wilcoxon rank-sum test (iii) in range $[1e-1,\dots,1e-20]$ (steps of e-1). All methods have large sub ranges of insignificantly well-ranking solutions for both Covering and F1 (data not shown). (i) performs best in range $[0.7,\dots,0.8]$, (ii) between $[0.025,\dots,0.035]$, and (iii) in range $[1e-5,\dots,1e-20]$. We set the default values to 0.75, 0.025, and 1e-15 for (i -- iii), respectively. We then compared (i -- iii) with each other to find the best method, see Figure~\ref{fig:cd_design_choices} (bottom right). All proposed methods rank insignificantly different. (iii) shows the lowest average Covering and F1 rank of 1.2 and highest average Covering (and F1) performance of 83.8\% (78.8\%). The gini gain and the ROC/AUC threshold are on par considering the average Covering/F1 ranks (1.3 -- 1.4) and performances. Thus, we choose (iii) for change point candidate validation.

\subsection{Segmentation Performance}
\label{sec:benchmark_segmentation}

We compare the performance of ClaSP with its six competitors in two different settings: With and without pre-defined $C$. Having no pre-defined number of segments is, by far, the hardest setting as here methods have to detect the amount of CPs as well as their locations. Note that not all methods support both modes. PELT can only learn the number of CPs automatically. We further partition the results per UTSA and TSSB benchmark, to discuss potential differences.

\paragraph{Segmentation with unknown number of Change Points}

\begin{figure}[t]
	\begin{minipage}{6cm}
        \includegraphics[width=1.0\columnwidth]{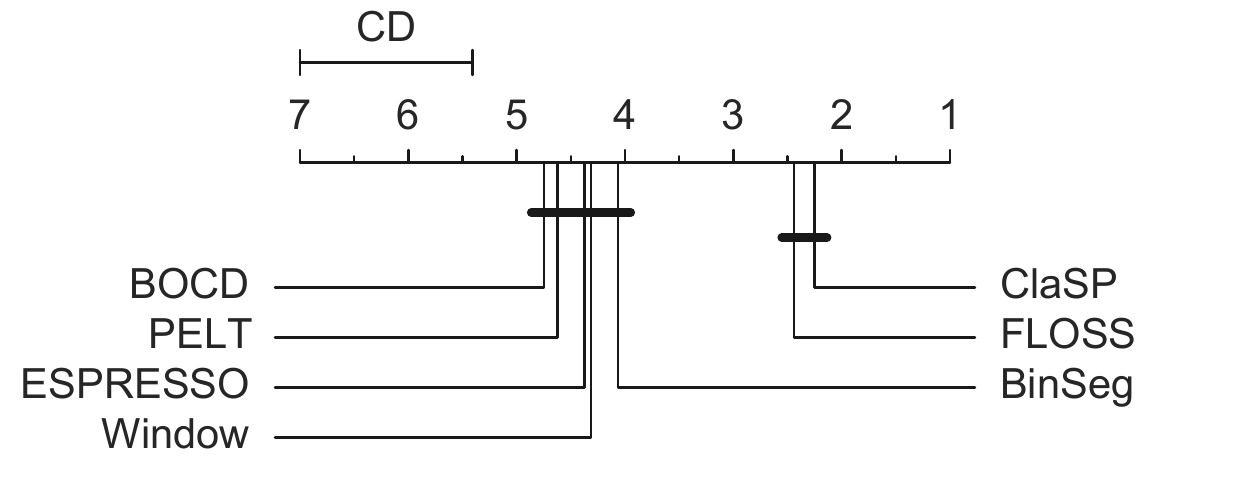}
	\end{minipage}
	\begin{minipage}{6cm}
        \includegraphics[width=1.0\columnwidth]{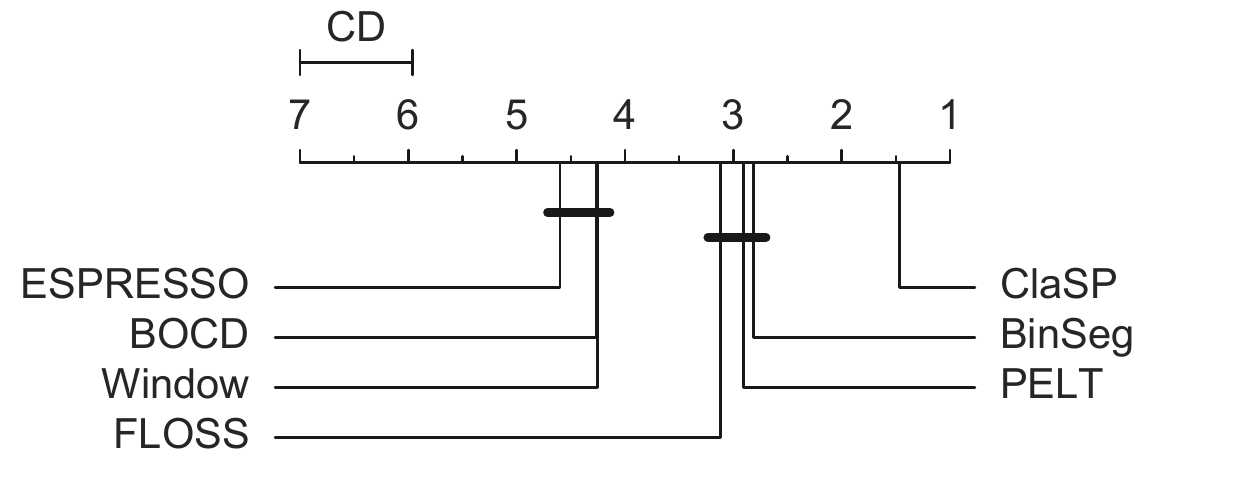}
	\end{minipage}
	\caption{Covering segmentation ranks on the $32$ UTSA (left) and $75$ TSSB (right) benchmark data sets for ClaSP (lowest rank) and the $6$ state-of-the-art competitors when the number of CPs is not known. For both ClaSP performs best.\label{fig:cd_unsupervised_seg}
	}
\end{figure}

\begin{figure}[t]
	\begin{minipage}{6cm}
        \includegraphics[width=1.0\columnwidth]{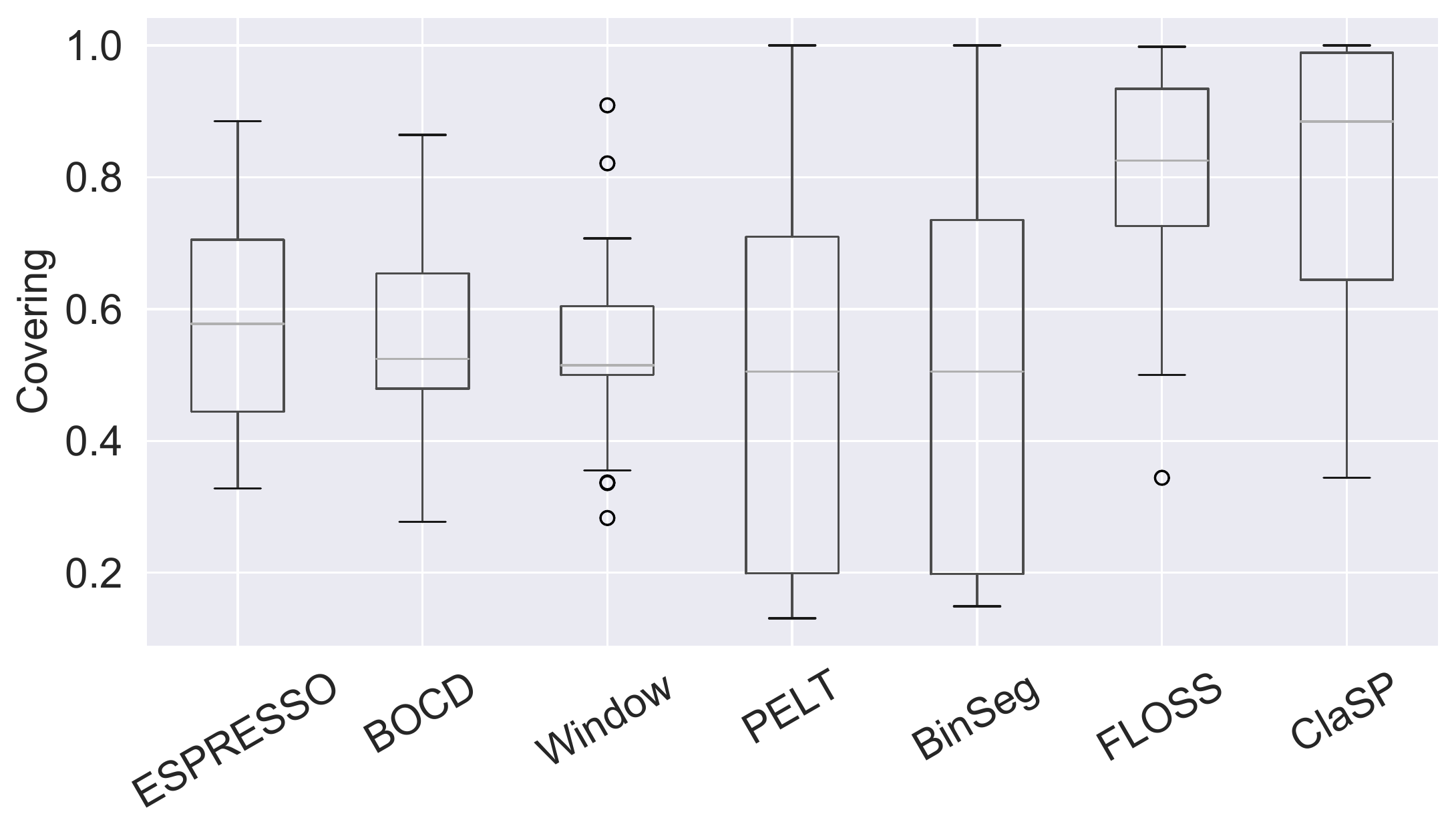}
	\end{minipage}
	\begin{minipage}{6cm}
        \includegraphics[width=1.0\columnwidth]{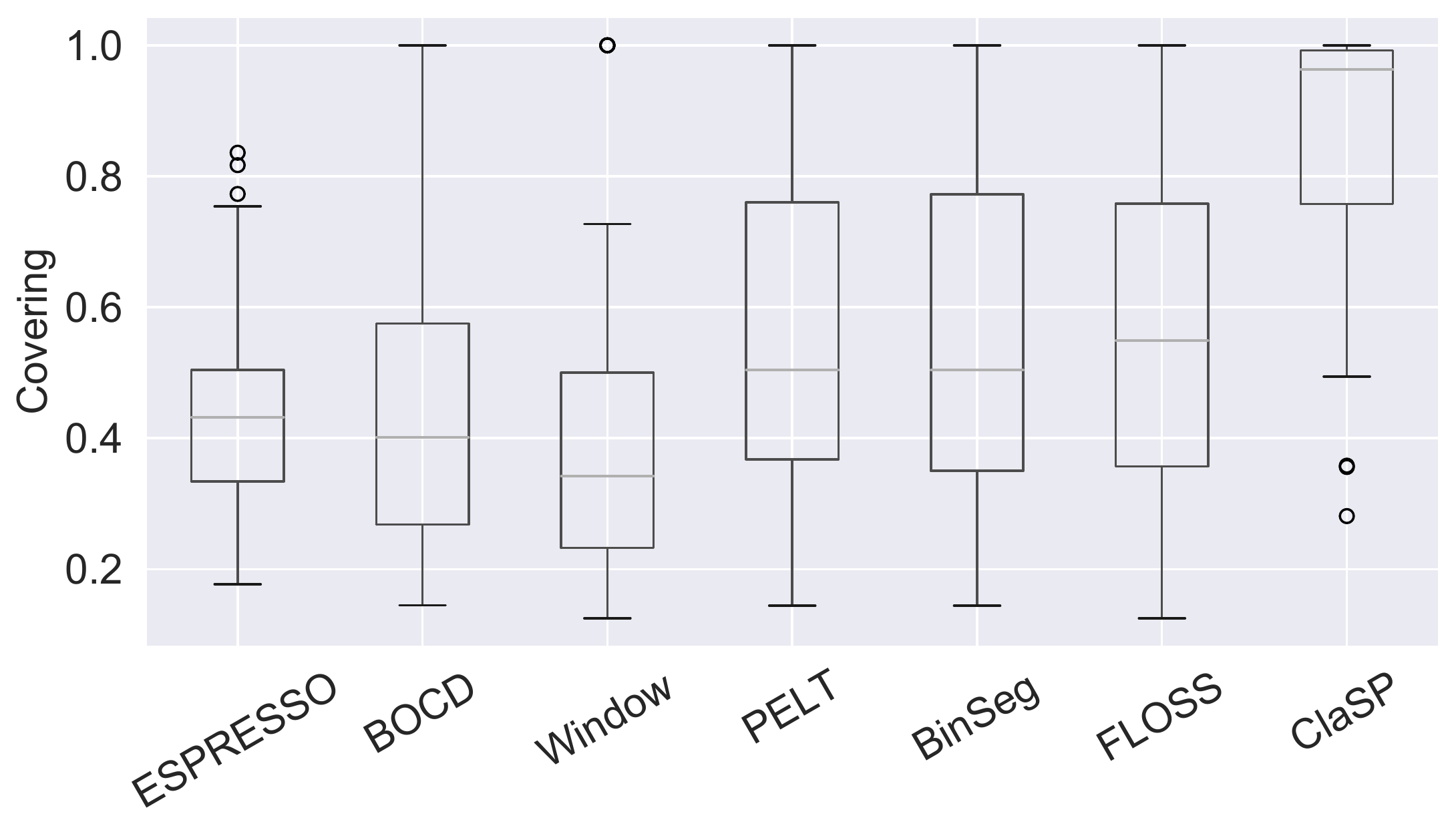}
	\end{minipage}
	\caption{Boxplot for Covering performance on the $32$ UTSA (left) and $75$ TSSB (right) benchmark data sets for ClaSP and the state-of-the-art competitors when the number of CPs is not known.\label{fig:boxplot_unsupervised_seg}
	}
\end{figure}

The critical difference diagrams in Figure~\ref{fig:cd_unsupervised_seg} show the average Covering ranks for ClaSP, FLOSS, PELT, BinSeg, Window, BOCD, and ESPRESSO based on their performances for the segmentation task without predefined number of segments per benchmark. For UTSA, ClaSP shows the best Covering rank (2.3), followed by FLOSS (2.5), BinSeg (4.1), Window (4.3), ESPRESSO (4.4), PELT (4.6), and BOCD (4.8). ClaSP and FLOSS are both significantly better than all other competitors. The F1 ranking is comparable, except that Window (2th rank) and PELT (5th rank) perform better. For TSSB, ClaSP ranks first (1.5), followed by BinSeg and PELT (2.8), FLOSS (3.1), Window (4.3), BOCD (4.3), and ESPRESSO (4.6). ClaSP significantly outperforms all other competitors. The F1 ranking is also similar, except that Window (4th rank) scores better results. The rankings show that ClaSP scores state-of-the-art results across benchmarks and evaluation metrics.

On all 107 data sets combined, ClaSP has $70$ wins or ties (first position in ranking), followed by FLOSS ($19$), BinSeg ($16$), PELT ($15$), Window ($8$), BOCD ($8$) and ESPRESSO ($7$) (counts do not sum up to $107$ due to ties) and is significantly more accurate than all other competitors. Interestingly, the $41$ TS for which ClaSP ranks 2nd position or worse, show no common similarities, like e.g. noticeable small data sets or TS with lots of CPs. ClaSP also ranks best for four subcases of TSS, namely with only TS that have no CP ($6$ instances), with just one CP ($46$ instances), with at least two CPs ($55$ instances), and with reoccurring sub-segments ($10$ instances).

\begin{table}[t]
	\caption{Summary wins/ties/losses of ClaSP over rivals when the number of CPs is not known.\label{tab:unsupervised_wtl}}	
	
	\begin{minipage}{12cm}
	    \captionof*{table}{UTSA}
        \begin{centering}
    		\begin{tabular}{c|cccccc}
    			\toprule 			
    			& FLOSS & BinSeg & PELT & Window & BOCD & ESPRESSO \tabularnewline
    			\hline 
    			ClaSP & $19/3/10$ & $23/3/6$ & $24/3/5$ & $25/1/6$ & $25/0/7$ & $26/0/6$ \tabularnewline
    			\bottomrule 			
    		\end{tabular}
    	\par\end{centering}    
	\end{minipage}

	\begin{minipage}{12cm}
        \captionof*{table}{TSSB}
        \begin{centering}
    		\begin{tabular}{c|cccccc}
    			\toprule 			
    			& FLOSS & BinSeg & PELT & Window & BOCD & ESPRESSO \tabularnewline
    			\hline 
    			ClaSP & $53/13/9$ & $56/12/7$ & $56/12/7$ & $60/15/0$ & $65/5/5$ & $68/0/7$ \tabularnewline
    			\bottomrule 			
    		\end{tabular}
	    \par\end{centering}
	\end{minipage}

\end{table}

\begin{table}[t]
	\caption{Mean, median, and standard deviation Covering performances for ClaSP and its competitors when the number of CPs is not known.\label{tab:unsupervised_summary}}

	\begin{minipage}{12cm}
	    \captionof*{table}{UTSA}
    	\begin{centering}
    		\begin{tabular}{c|ccccccc}
    			\toprule 			
    			Covering & ClaSP & FLOSS & BinSeg & PELT & Window & BOCD & ESPRESSO\tabularnewline
    			\hline 
    			 mean & $\textbf{79.8\%}$ & $79.0\%$ & $52.7\%$ & $50.4\%$ & $53.8\%$ & $55.5\%$ & $58.0\%$ \tabularnewline
    			 median & $\textbf{88.5\%}$ & $82.5\%$ & $50.5\%$ & $50.5\%$ & $51.4\%$ & $52.4\%$ & $57.7\%$ \tabularnewline
    			 std & $20.4\%$ & $17.2\%$ & $30.6\%$ & $30.0\%$ & $\textbf{12.9\%}$ & $14.4\%$ & $15.8\%$ \tabularnewline
    			\bottomrule 			
    		\end{tabular}
    	\par\end{centering}
	\end{minipage}

	\begin{minipage}{12cm}
        \captionof*{table}{TSSB}
    	\begin{centering}
    		\begin{tabular}{c|ccccccc}
    			\toprule 			
    			Covering & ClaSP & FLOSS & BinSeg & PELT & Window & BOCD & ESPRESSO\tabularnewline
    			\hline 
    			 mean & $\textbf{85.5\%}$ & $56.7\%$ & $57.5\%$ & $58.1\%$ & $40.1\%$ & $44.9\%$ & $44.4\%$ \tabularnewline
    			 median & $\textbf{96.3\%}$ & $54.9\%$ & $50.4\%$ & $50.4\%$ & $34.2\%$ & $40.1\%$ & $43.2\%$ \tabularnewline
    			 std & $18.8\%$ & $25.2\%$ & $25.6\%$ & $24.7\%$ & $22.9\%$ & $20.3\%$ & $\textbf{15.5\%}$ \tabularnewline
    			\bottomrule 			
    		\end{tabular}
	    \par\end{centering}
	\end{minipage}
	
\end{table}

It further scores the highest mean/median Covering performance across benchmarks (Table~\ref{tab:unsupervised_summary}, Figure~\ref{fig:boxplot_unsupervised_seg}). For UTSA, the summary statistics of ClaSP to its second-best competitor FLOSS are comparable and confirm the ranking. For TSSB, however, ClaSP has a $28.0$ percentage points (pp) higher average Covering performance and $6.8$ pp smaller standard deviation compared to its second-best competitor BinSeg. In a pairwise comparison of ClaSP against every competitor on UTSA (TSSB), ClaSP achieves between $19$ ($53$) wins (vs FLOSS) and $26$ ($68$) wins (vs ESPRESSO) (Table~\ref{tab:unsupervised_wtl}). For instance, it achieves a higher Covering score than FLOSS for $19$ ($53$) of the $32$ ($75$) UTSA (TSSB) data sets, the same performance in $3$ ($13$) cases, and is beaten by FLOSS also in $10$ ($9$) cases. Based on this analysis, we find that ClaSP is able to successfully learn appropriate values for its model-parameters and significantly outperform state-of-the-art competitors in the general segmentation task on the considered benchmark data sets. Interestingly, ClaSP's performance differences are more pronounced for the TSSB benchmark when compared to UTSA. This may be because TSSB contains more difficult problem settings (e.g. more and reoccurring segments, more signal types), when compared to UTSA, that ClaSP segments more accurately due to its score profile,  advanced segmentation procedure and data-dependent model-parameter selection. FLOSS, in comparison, uses the pre-defined window size as a hyper-parameter and greedily determines all valleys in its arc curve that are smaller or equal to a determined threshold (for all data sets). For complex scenarios, as e.g. in Figure \ref{fig:ClaSP_ensembling}, this can lead to many bad decisions that are penalized in the comparative analysis. In this instance, ClaSP achieves $98.7\%$ Covering while FLOSS scores only $37.7\%$.

\paragraph{Segmentation with known number of Change Points}

\begin{figure}[t]
	\begin{minipage}{6cm}
        \includegraphics[width=1.0\columnwidth]{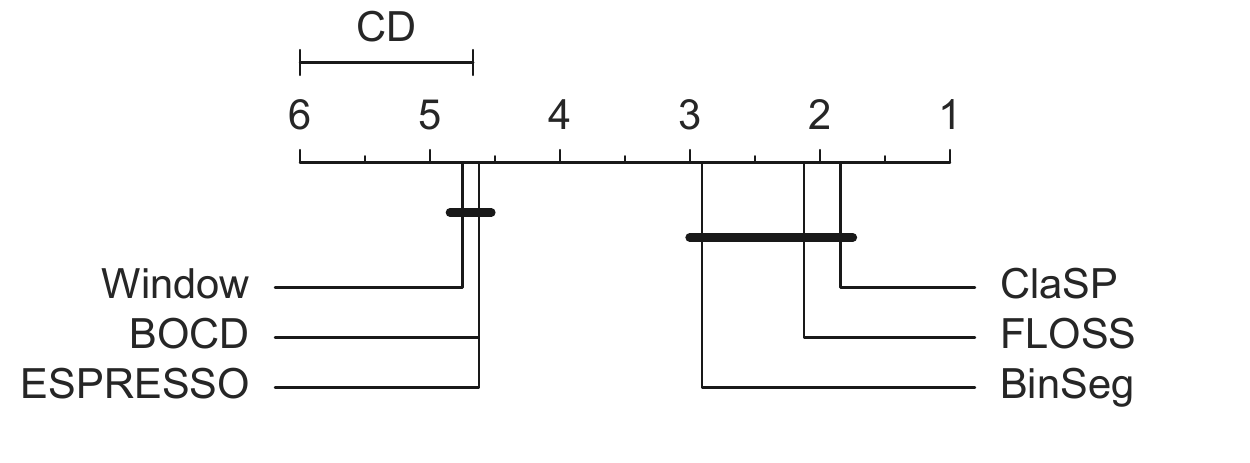}
	\end{minipage}
	\begin{minipage}{6cm}
        \includegraphics[width=1.0\columnwidth]{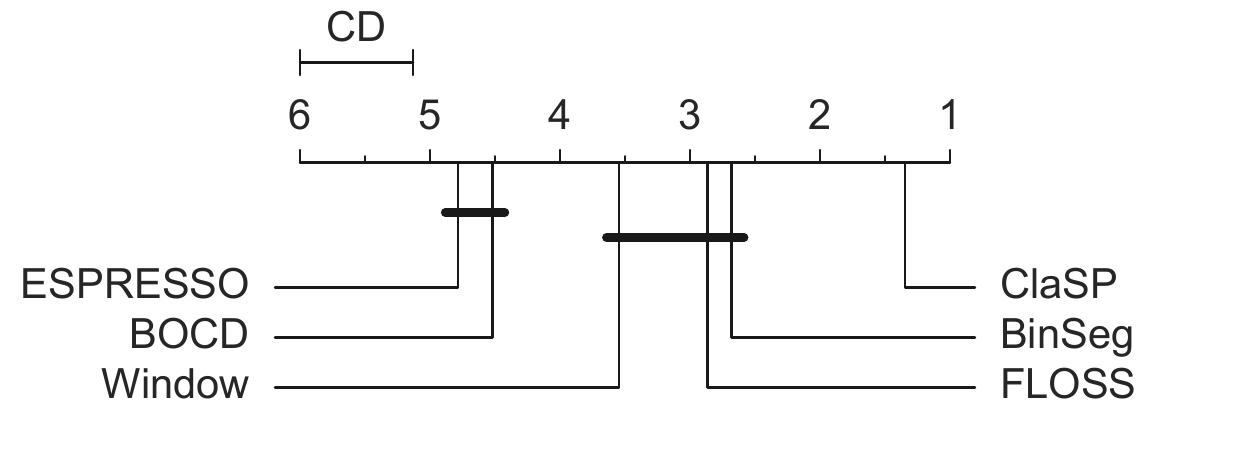}
	\end{minipage}
	\caption{Covering segmentation ranks on the $32$ UTSA (left) and $75$ TSSB (right) benchmark data sets for ClaSP (lowest rank) and $5$ state-of-the-art competitors with known number of CPs.\label{fig:cd_semi-supervised_seg}
	}
\end{figure}

\begin{figure}[t]
	\begin{minipage}{6cm}
        \includegraphics[width=1.0\columnwidth]{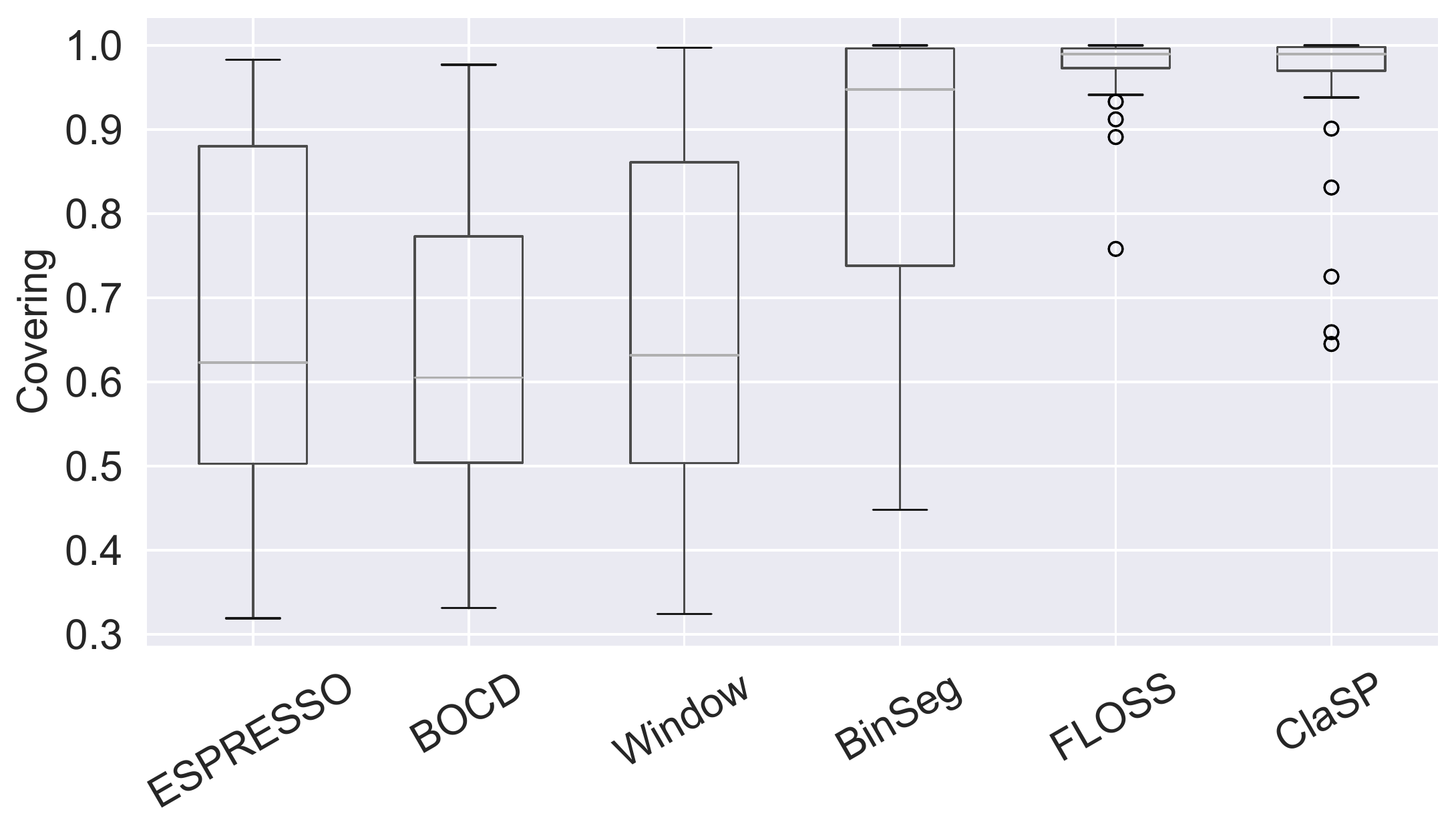}
	\end{minipage}
	\begin{minipage}{6cm}
        \includegraphics[width=1.0\columnwidth]{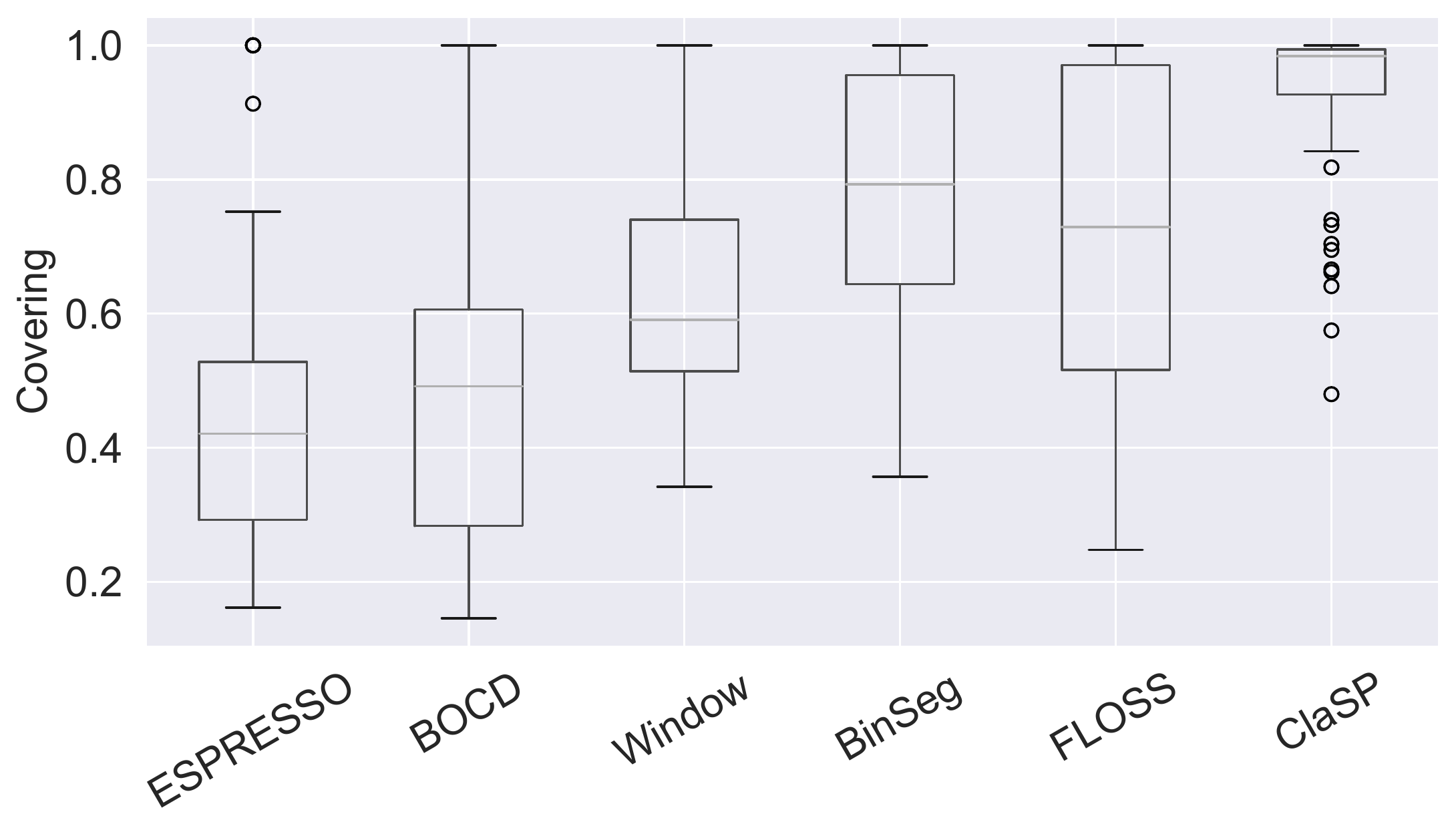}
	\end{minipage}
	\caption{Boxplot for Covering performance on the $32$ UTSA (left) and $75$ TSSB (right) benchmark data sets for ClaSP and the state-of-the-art competitors with known number of CPs.\label{fig:boxplot_semi-supervised_seg}
	}
\end{figure}

Figure~\ref{fig:cd_semi-supervised_seg} contains the average Covering ranks for all methods except PELT, which cannot set the amount of segments as a parameter, for the segmentation task when the number of CPs is predefined. The rank order and the number of wins or ties per data set is comparable to the previous setting. For UTSA, ClaSP (1.8) on average ranks better than all other competitors, followed by FLOSS (2.1), BinSeg (2.9), ESPRESSO and BOCD (4.6), as well as Window (4.8). ClaSP, FLOSS and BinSeg significantly outperform all other competitors. The F1 ranking is similar, except that FLOSS (1st rank) and Window (4th rank) perform better and the three best-ranking methods tie on most data sets. For TSSB, ClaSP (1.3) ranks first, followed by BinSeg (2.7), FLOSS (2.9), Window (3.5), BOCD (4.5), and ESPRESSO (4.8). ClaSP is significantly more accurate than its competitors. The F1 ranking order is identical. If we consider the aforementioned subtasks of segmentation on all $107$ data sets combined, all methods rank equally well for TS without CPs; in all other cases, ClaSP beats all competitors. This shows that ClaSP also scores state-of-the-art performances across benchmarks and evaluation metrics for the constrained segmentation task, where the number of CPs is known.

\begin{table}[t]
	\caption{Summary wins/ties/losses of ClaSP over rivals with known number of CPs.\label{tab:semi-supervised_wtl}}	
	\begin{minipage}{12cm}
	\captionof*{table}{UTSA}
	\begin{centering}
		\begin{tabular}{c|cccccc}
			\toprule 			
			& FLOSS & BinSeg & Window & BOCD & ESPRESSO \tabularnewline
			\hline 
			ClaSP & $19/2/11$ & $23/0/9$ & $31/0/1$ & $29/0/3$ & $29/0/3$ \tabularnewline
			\bottomrule 			
		\end{tabular}
	\par\end{centering}
	\end{minipage}
	\begin{minipage}{12cm}
	\captionof*{table}{TSSB}
	\begin{centering}
		\begin{tabular}{c|cccccc}
			\toprule 			
			& FLOSS & BinSeg & Window & BOCD & ESPRESSO \tabularnewline
			\hline 
			ClaSP & $59/7/9$ & $57/7/11$ & $67/6/2$ & $66/6/3$ & $68/6/1$ \tabularnewline
			\bottomrule 			
		\end{tabular}
	\par\end{centering}
	\end{minipage}
\end{table}

\begin{table}[t]
	\caption{Mean, median, and standard deviation Covering performances for ClaSP and its competitors with known number of CPs. \label{tab:semi-supervised_summary}}
	\begin{minipage}{12cm}
	\captionof*{table}{UTSA}
	\begin{centering}
		\begin{tabular}{c|ccccccc}
			\toprule 			
			Covering & ClaSP & FLOSS & BinSeg & Window & BOCD & ESPRESSO\tabularnewline
			\hline 
			 mean & $95.1\%$ & $\textbf{97.2\%}$ & $84.5\%$ & $67.2\%$ & $63.7\%$ & $67.0\%$ \tabularnewline
			 median & $\textbf{99.0\%}$ & $\textbf{99.0\%}$ & $94.8\%$ &$63.2\%$ & $60.5\%$ & $62.3\%$ \tabularnewline
			 std & $9.7\%$ & $\textbf{4.7\%}$ & $20.2\%$ & $20.9\%$ & $18.5\%$ & $21.0\%$ \tabularnewline
			\bottomrule 			
		\end{tabular}
	\par\end{centering}
	\end{minipage}
	\begin{minipage}{12cm}
	\captionof*{table}{TSSB}
	\begin{centering}
		\begin{tabular}{c|ccccccc}
			\toprule 			
			Covering & ClaSP & FLOSS & BinSeg & Window & BOCD & ESPRESSO\tabularnewline
			\hline 
			 mean & $\textbf{93.1\%}$ & $71.8\%$ & $77.5\%$ & $63.9\%$ & $49.4\%$ & $46.1\%$ \tabularnewline
			 median & $\textbf{98.4\%}$ & $72.9\%$ & $79.3\%$ &$59.1\%$ & $49.2\%$ & $42.1\%$ \tabularnewline
			 std & $\textbf{11.3\%}$ & $23.9\%$ & $19.0\%$ & $17.3\%$ & $24.1\%$ & $22.1\%$ \tabularnewline
			\bottomrule 			
		\end{tabular}
	\par\end{centering}
	\end{minipage}
\end{table}

Overall, for UTSA ClaSP has the second highest (yet comparable) Covering summary statistics, compared to FLOSS, and for TSSB it has the highest mean and median Covering performance and by far the lowest standard deviation (Table~\ref{tab:semi-supervised_summary}, Figure~\ref{fig:boxplot_semi-supervised_seg}). On average for UTSA (TSSB), it scores $15.3$ ($7.6$) pp higher compared to the (harder) problem with unknown $C$. This shows that ClaSP is able to detect the correct amount of CPs with high accuracy if needed. FLOSS, in comparison, improves $18.2$ ($15.1$) pp and BinSeg $31.8$ ($20.0$) pp. In a pairwise comparison of ClaSP against every competitor on UTSA (TSSB), ClaSP achieves between $19$ ($57$) wins (vs FLOSS or rather BinSeg) and $31$ ($68$) wins (vs Window or ESPRESSO) (Table~\ref{tab:semi-supervised_wtl}). For instance, it achieves a higher Covering score than FLOSS for $19$ ($59$) of the $32$ ($75$) UTSA (TSSB) data sets, the same performance in $2$ ($7$) cases, and is beaten by FLOSS in $11$ ($9$) cases. From this evaluation, we conclude that ClaSP is also able to significantly outperform state-of-the-art competitors on the considered data sets in the constraint segmentation task, when $C$ is known a priori, making it a viable alternative to existing TSS technology. As expected, all methods improve their performances, on average, when the number of CPs is known, as this simplifies the segmentation problem. Interestingly, ClaSP ranks 1st place across problem settings and benchmarks, which emphasizes its robustness and quality of computation. If we revisit the complex scenario in Figure \ref{fig:ClaSP_ensembling}, for example, ClaSP scores $98.7\%$ Covering, independent of it having to learn $C$ or not, while FLOSS only improves $2.2$ pp (from $37.7\%$ to $39.9\%$).

\subsection{Runtime Comparison} \label{sec:runtime}

\begin{figure}[t]
	\includegraphics[width=1.0\textwidth]{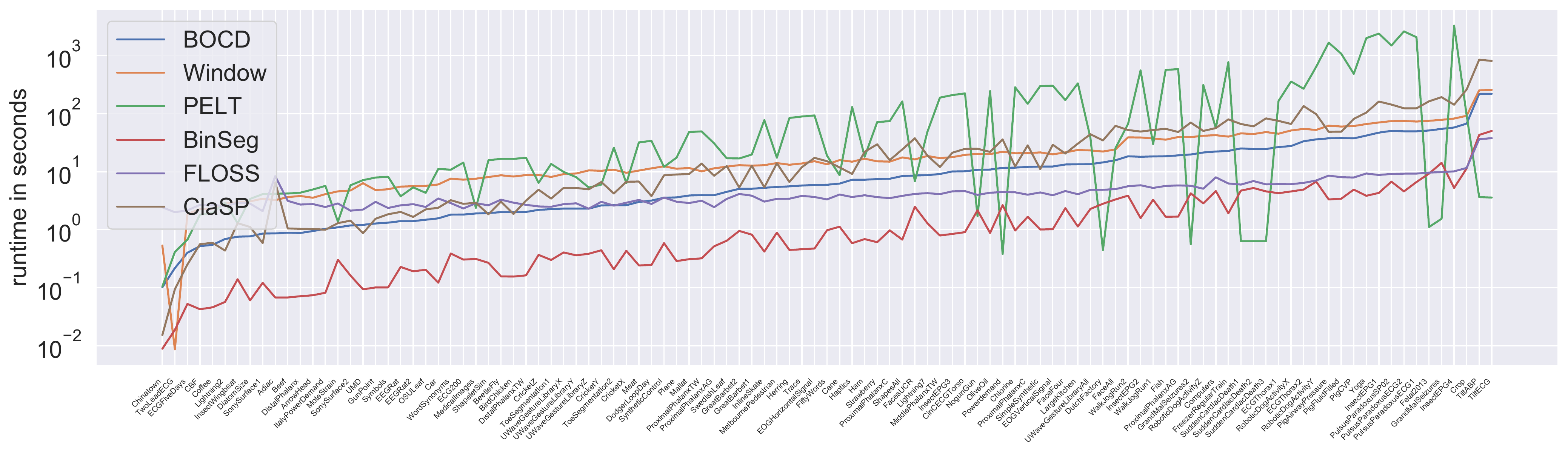}
	\caption{Runtime of ClaSP in comparison to its competitors. Data sets are ordered by size with highly irregular increases.\label{fig:runtime}}
\end{figure}

\begin{figure}[t]
	\begin{minipage}{6cm}
        \includegraphics[width=1.0\columnwidth]{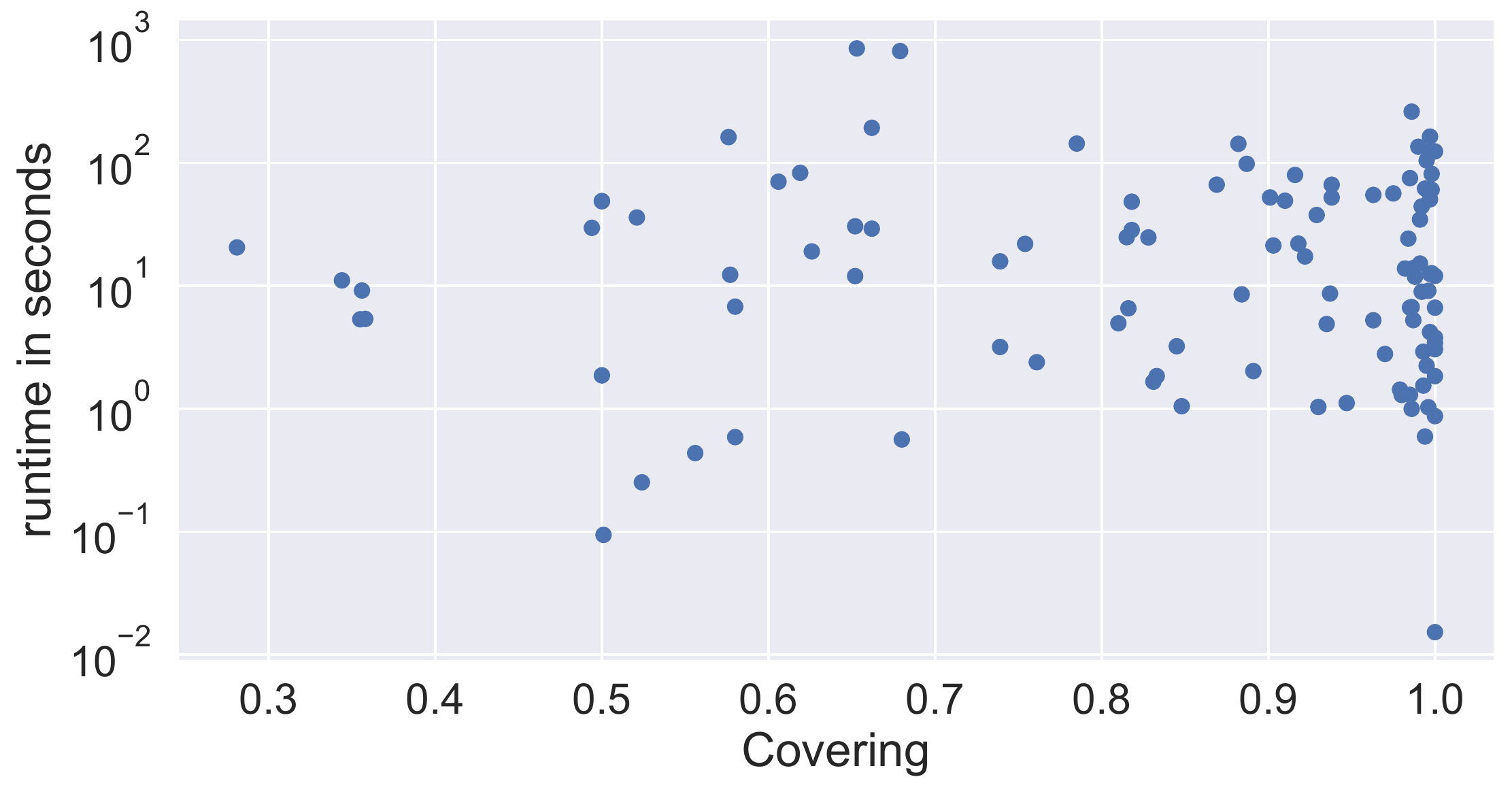}
	\end{minipage}
	\begin{minipage}{6cm}
        \includegraphics[width=1.0\columnwidth]{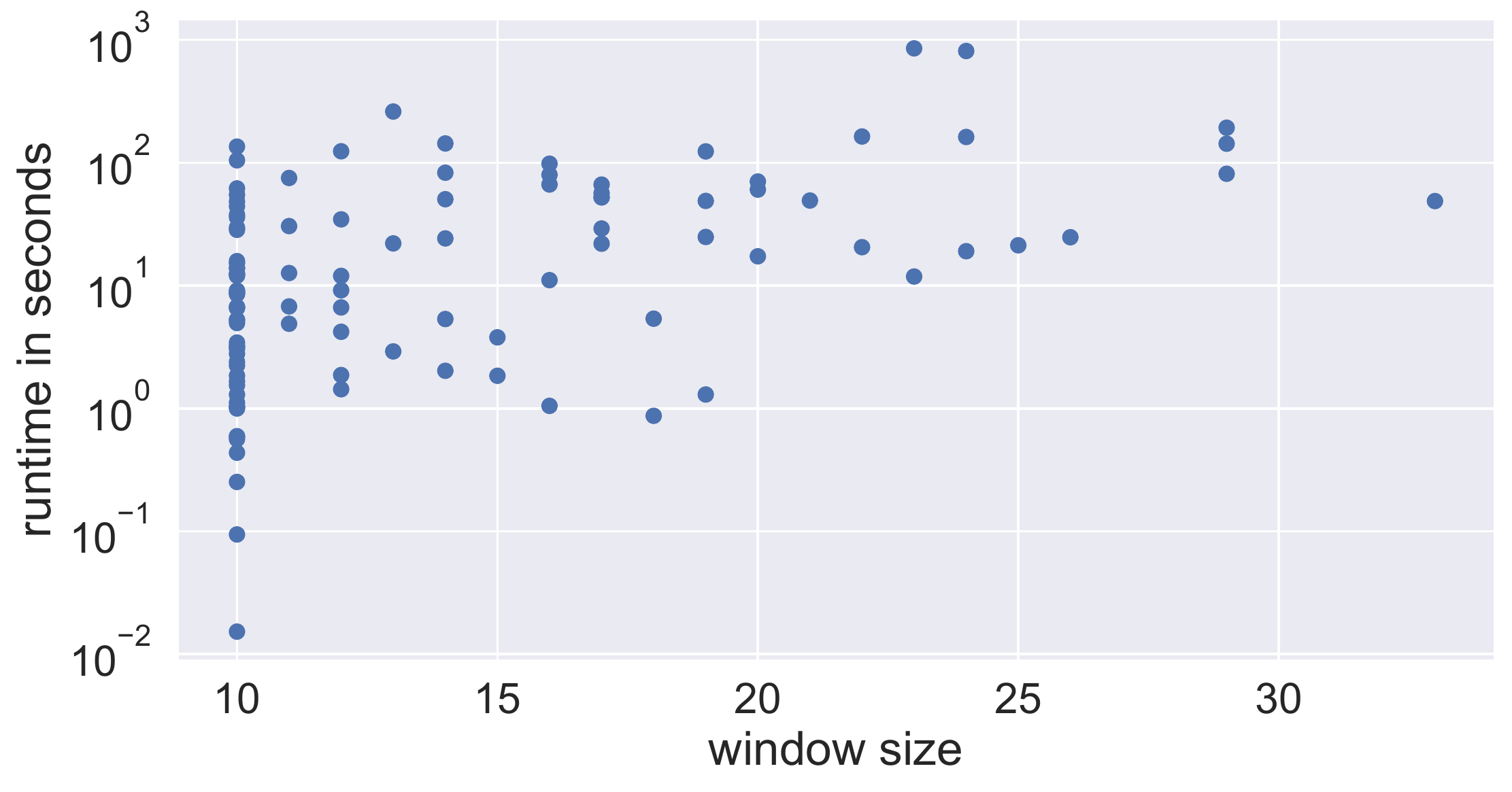}
	\end{minipage}
	\begin{minipage}{6cm}
        \includegraphics[width=1.0\columnwidth]{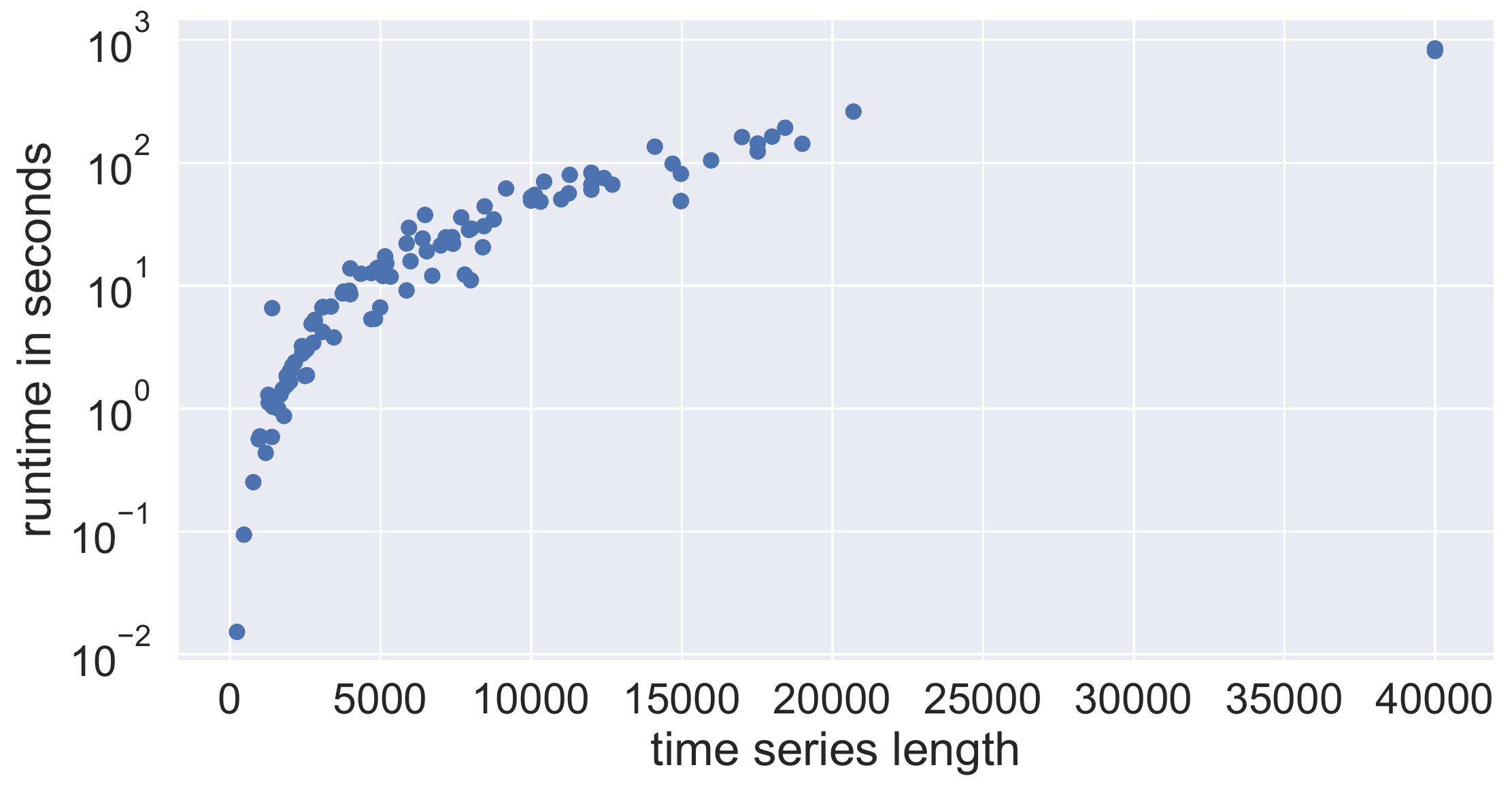}
	\end{minipage}
	\begin{minipage}{6cm}
        \includegraphics[width=1.0\columnwidth]{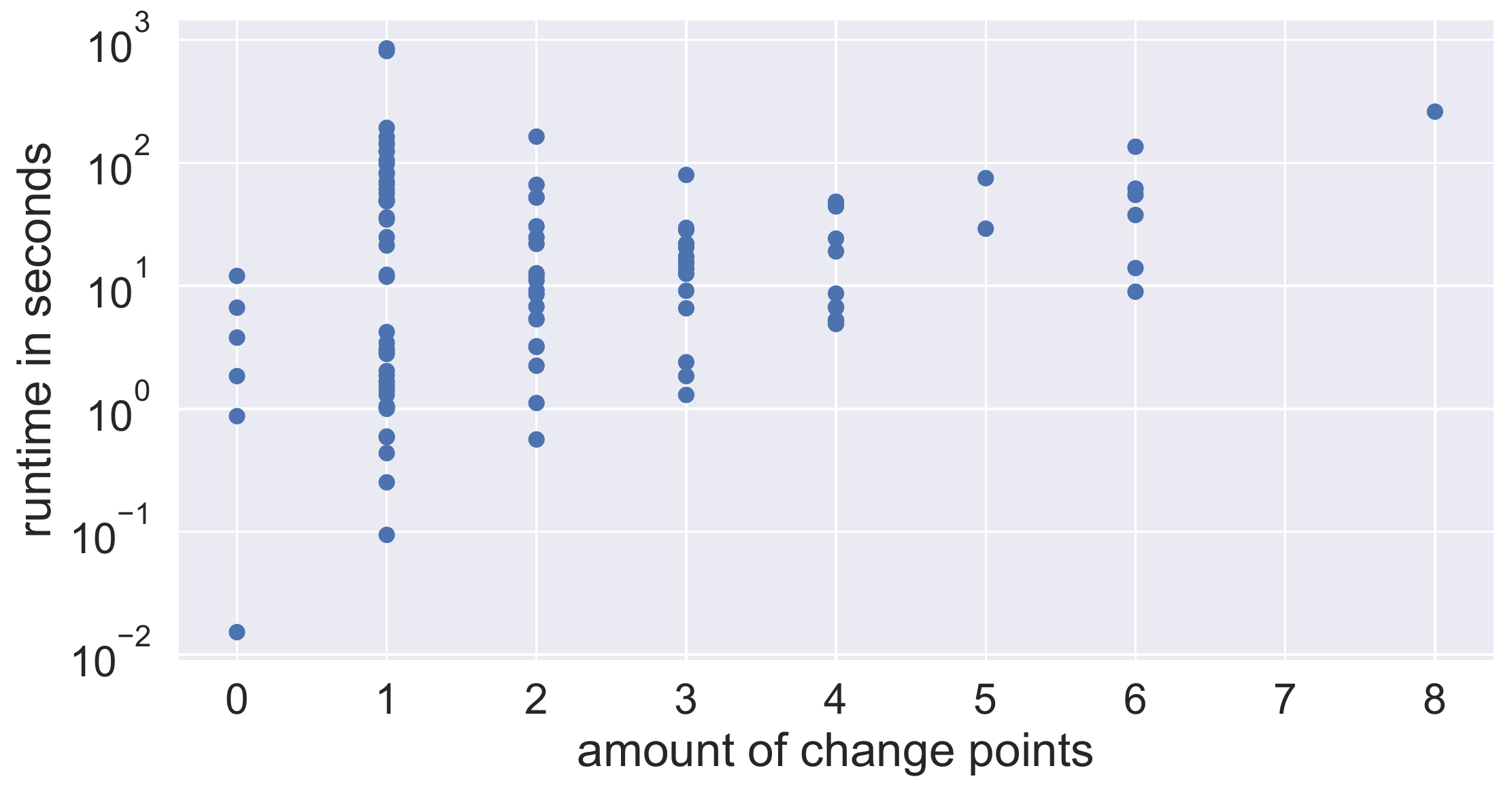}
	\end{minipage}
	\caption{Scalability of ClaSP considering Covering performance (top left), window size (top right), time series length (bottom left), and number of CPs (bottom right). \label{fig:scalability}
	}
\end{figure}

Figure~\ref{fig:runtime} shows runtimes of ClaSP, FLOSS, PELT, BinSeg, Window, BOCD on each of the $107$ benchmark TS. On average, BinSeg has the lowest overall runtime (2.6s), followed by FLOSS (5.1s), BOCD (16.6s), Window (27.5s) ClaSP (48.6s), and PELT (239.2s). Although FLOSS and ClaSP in principle have the same complexity of $\mathcal{O}(n^2)$, ClaSP is substantially slower due to its more elaborate segmentation procedure that runs ins $\mathcal{O}(C \cdot n^2)$. Nevertheless, ClaSP is faster than PELT (for $92$ data sets) and Window ($62$ data sets). The marked peak in runtime for TiltABP and TiltECG are due to their much greater length (length 40k compared to length 19k of the next-longest TS InsectEPG4). 

Scalability results relative to Covering performance, window size, time series length, and amount of CPs are shown in Figure~\ref{fig:scalability}. We observe no correlation between runtime and Covering performance or window sizes. On the other hand, ClaSP becomes slower with increasing TS length or increasing numbers of CPs. It segments most TS with less than 5k data points in under 10 seconds, but needs up to 100 seconds for TS with 5k to 10k observations (which roughly follows the expected quadratic course). We consider faster approximate versions of ClaSP as part of our future work.

\subsection{Complex Classifiers in ClaSP} \label{sec:complex-classifiers}

\begin{figure}[t]
	\includegraphics[width=1.0\columnwidth]{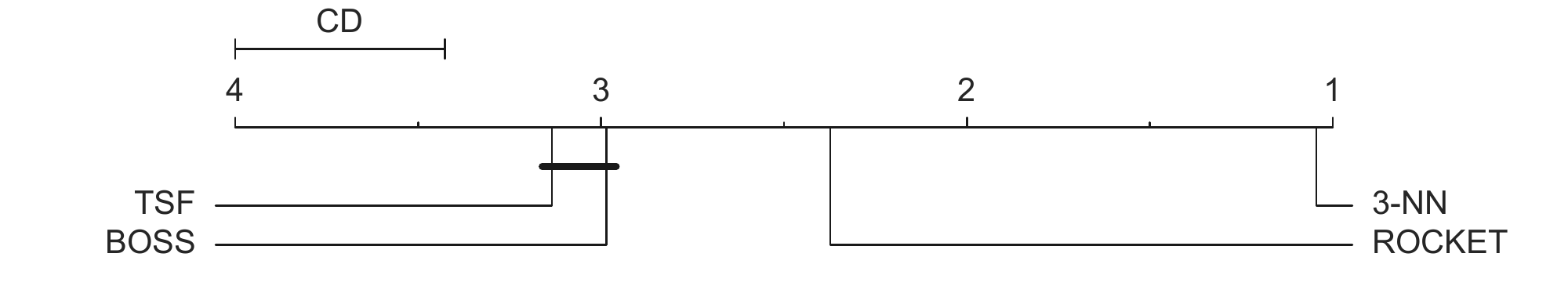}
	\caption{Covering segmentation ranks on $67$/$107$ benchmark data sets for ClaSP with 3-NN (lowest rank) and other classifiers with known number of CPs.\label{fig:cd_clf}
	}
\end{figure}

All results so-far were achieved with a 3-NN classifier for the evaluation of split points. However, we also performed initial experiments with more sophisticated TS classifiers, including BOSS~\citep{schafer2014boss}, TS-Forest~\citep{deng2013time} and ROCKET~\citep{dempster2020rocket}. We evaluated all complex classifiers in ClaSP using a 5-fold cross-validation with ROC/AUC score and choose to report results for $67$/$107$ benchmark TS (2-7 segments, no reoccurring sub-segments), with less than 10.1k data points, as the computational load is impractical for larger time series given the scope of our analysis. Figure~\ref{fig:cd_clf} contains the average Covering ranks for all classifiers for the segmentation task with the number of CPs known. The 3-NN shows by far the best Covering rank (1.0), followed by ROCKET (2.4), BOSS (3.0), and TSF (3.1) and is significantly better than the complex classifiers. The 3-NN has $64$ wins or ties on the entire data set, followed by ROCKET ($4$). BOSS and TSF score no wins at all. The F1 ranking is similar, except that ROCKET only ranks insignificantly better than TSF and BOSS, which change third and fourth place. The 3-NN also has the highest mean (97.0\%) and median (98.7\%) Covering performance and by far the lowest standard deviation (5.2\%). The unoptimized 5-fold cross-validation also results in much higher runtimes, orders of magnitude slower than the 3-NN. Nonetheless, we consider optimizing complex classifiers for ClaSP as our future work.

\subsection{Results on Specific Use Cases}\label{sec:use_cases}

We describe results on three particularly interesting TS in more detail to show the strengths and limitations of ClaSP. All TS are part of the benchmarks used in the previous sections. 

\subsubsection*{Human Activity Recognition}

\begin{figure}[t]
	\includegraphics[width=1.0\columnwidth]{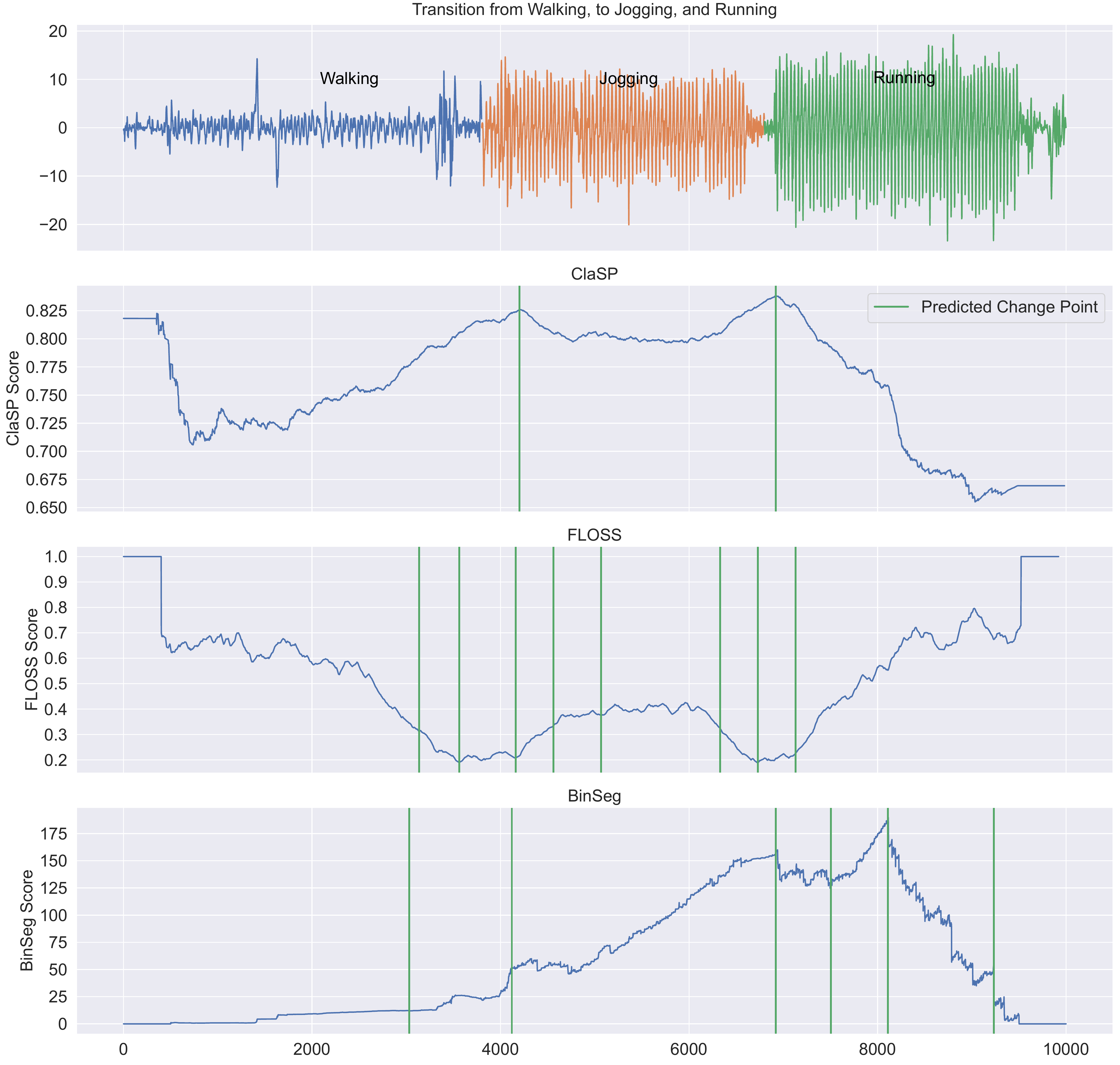}
	\caption{Segmentation of walking motions. There are two annotated change points, indicated by the transition in colours at which the activity types change.
	\label{fig:har}
	}
\end{figure}

Monitoring human behavior can reveal important insights about activity types and may help to improve health status or activity-aware services~\citep{Feuz2015AutomatedDO}. Figure~\ref{fig:har} shows the rotation of a subject's left calf while first walking, then jogging, and finally running~\citep{Baos2014DealingWT}. The TS captures the y-axis measurements from a gyroscope. Each type of activity represents one rather homogeneous segment, whereas different activities exhibit different frequencies of rotational deflections. The TS in the top of the figure contains two annotated change points, indicated by the transition in colors. From top to bottom, we highlight the profiles and segmentations given by the three best-ranking methods ClaSP, FLOSS and BinSeg. ClaSP has a smooth score profile with two large local maxima, that correspond to the activity transitions in the data set. It infers the correct amount of CPs (2) and their locations with only small deviations to the ground truth. FLOSS's arc curve is very noisy with many local deflections (valleys) in its profile. This makes extraction of CPs hard, and FLOSS  greedily extracts too many CPs in close proximity to each other. BinSeg wrongly identifies the center of the third segment as the most dominant change, but correctly locates the two activity changes as local maxima in its cost function. Its profile, however, has many sharp deflections and is not as interpretable as the two others. This use case highlights that ClaSP is able to detect the amount of CPs as well as their location with great accuracy without any pre-defined domain knowledge in the form of hyper-parameters.

\subsubsection*{Traffic Volume without Change Points}

\begin{figure}[t]
	\includegraphics[width=1.0\columnwidth]{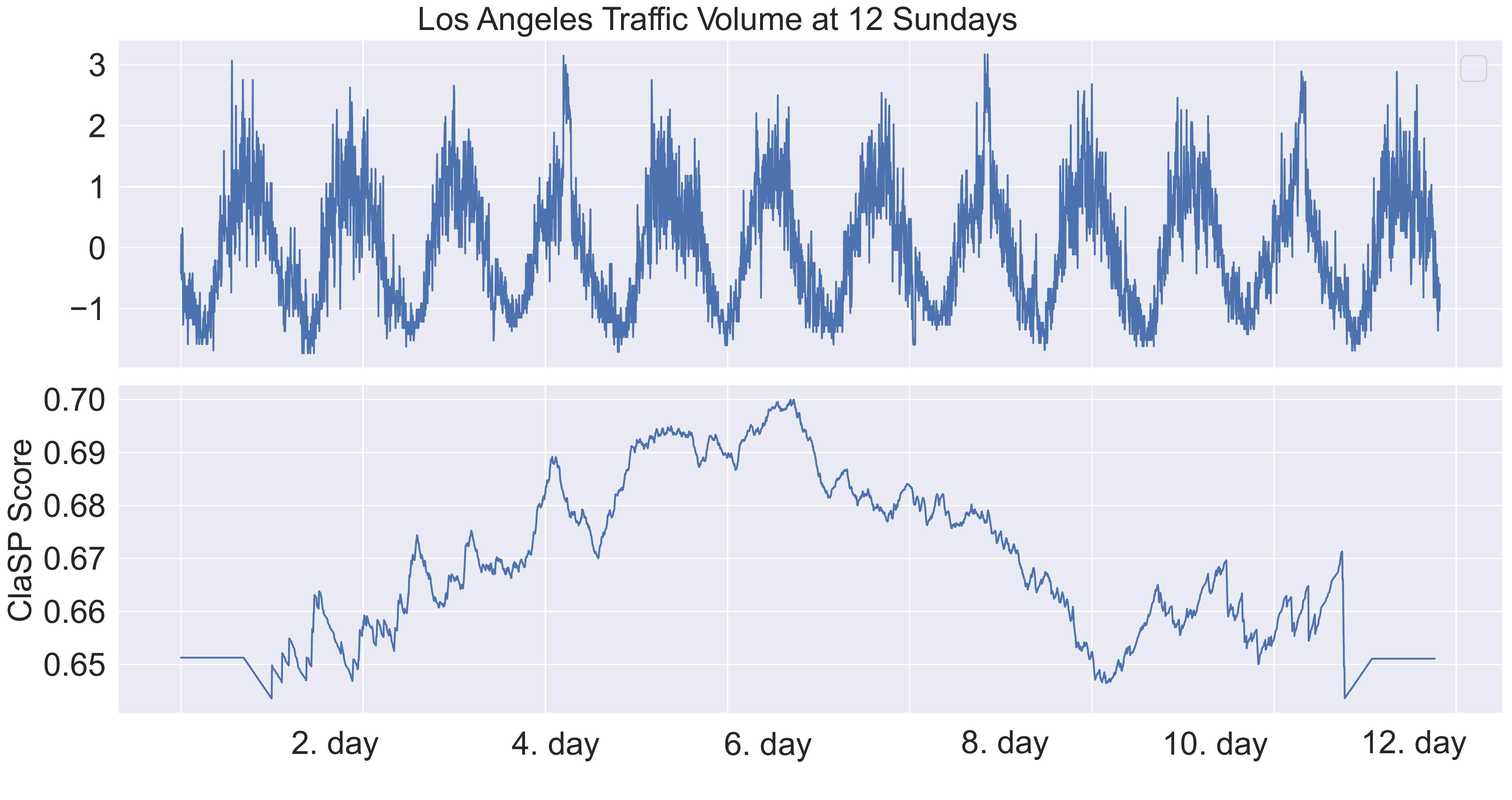}
	\caption{Los Angeles traffic volume at 12 Sundays near the Dodgers Stadium (top) with the corresponding ClaSP. No CPs are present (nor detected by ClaSP) in the TS.\label{fig:dodger-loop-day}
	}
\end{figure}

One interesting, yet underreported, special case in TSS are TS without CPs. A well-performing segmentation algorithm should be conservative and confident in its change detection, as CPs are inherently rare compared to the TS length. As an example, we show the ClaSP for a TS that contains the traffic volume for the 101 North freeway in Los Angeles observed at 12 Sundays~\citep{Ihler2006AdaptiveED} in Figure~\ref{fig:dodger-loop-day}. Traffic patterns are typically similar for a particular weekday (and distinct for different ones). The illustrated TS therefore reflects no change and just one segment. ClaSP only shows small deflections in range of $5$ pp and reports no CP (just as FLOSS and BinSeg). This shows that ClaSP can correctly detect the absence of change when the data shows homogenous behavior, which is a typical scenario in real-world applications.

\subsubsection*{Gradual vs Discrete Changes}

\begin{figure}[t]
	\includegraphics[width=1.0\columnwidth]{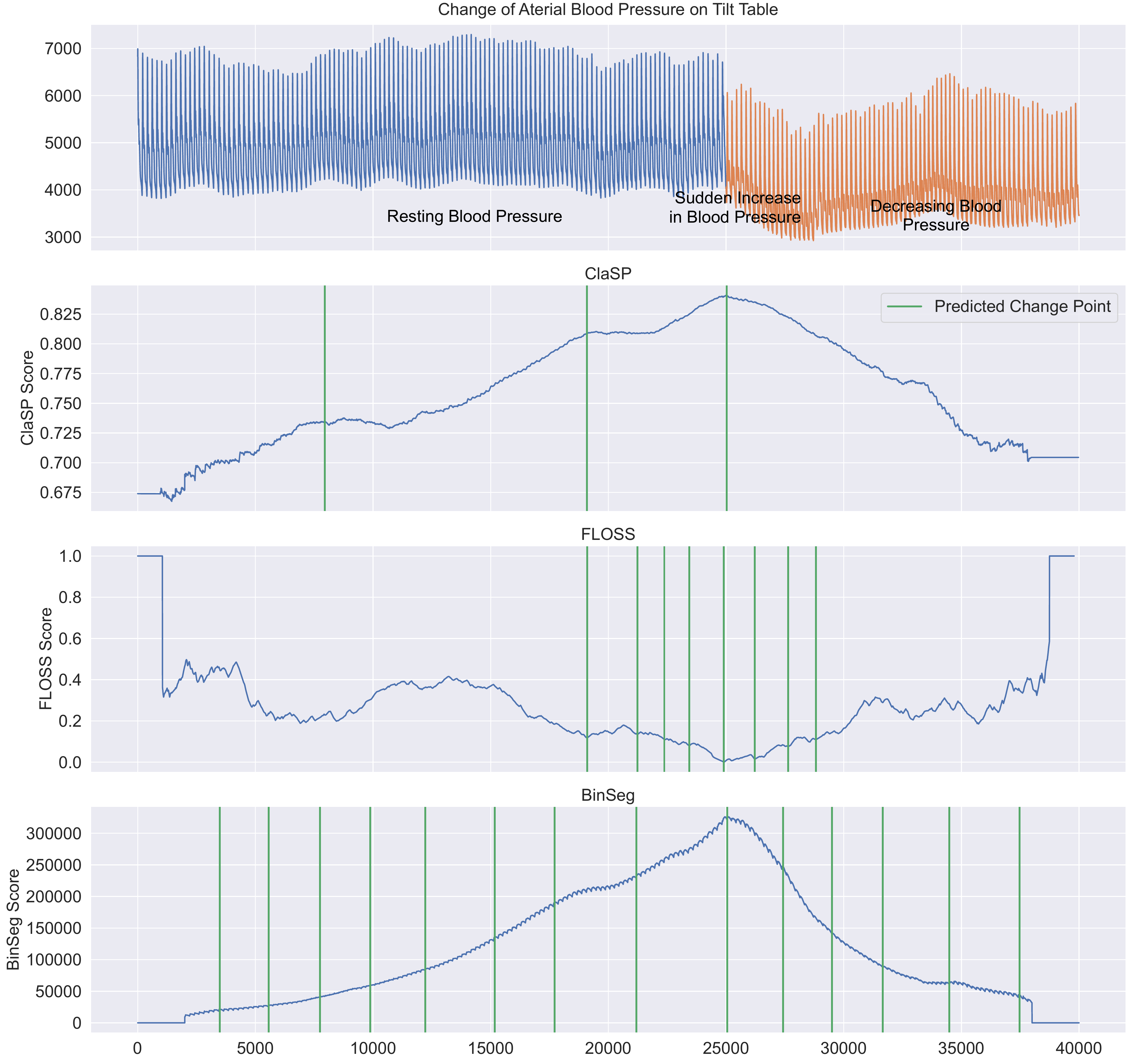}
	\caption{Segmentation of arterial blood pressure from a subject laying on a tilt table. The CP at 2.5s represents the start of the tilt to stand-up position.\label{fig:tilt-abp}
	}
\end{figure}

We study a use case showing the limits of TSS algorithms for cases where changes are gradual and not sudden. We analyze TS data from an experiment where the arterial blood pressure of a person lying on a tilt table with foot rest is measured~\citep{heldt2003circulatory}. The table is rapidly turned up, leading to a sudden rise in blood pressure. However, once reaching the upright position, the blood pressure only slowly goes back to normal. Figure~\ref{fig:tilt-abp} shows that ClaSP (top) detects the point in time when the table started tilting, i.e., it detects the rapid change, as well as two other non-annotated locations, as substantial deflections in its profile. FLOSS (center) reports multiple changes from its arc curve right before, during, and after the tilt happens. Its profile is noisy and contains many small local minima and maxima, which lead to a cluster of false positives around the true CP in the extraction process. BinSeg (bottom) detects a flood of splits throughout the entire TS. Its cost function is comparable to ClaSP, but the determined CP penalty (for all data sets) does not seem to be a good fit for this scenario. This use case underlines that automatic parameter selection for segmentation algorithms is non-trivial and requires either domain knowledge or advanced strategies, as discussed in this work. Further, none of the procedures detects the point where the upright position is reached, and blood pressure starts to decrease, as this does not lead to sudden changes in the TS, but happens only gradually. The detection of such complex phenomena would require a segmentation algorithm to detect abrupt shifts as well as trends, which is actually outside the scope of current TSS methods.

\section{Conclusion}\label{sec:conclusion}

We have introduced ClaSP, a new parameter-free method for TSS based on the principle of self-supervision. ClaSP produces and analyses a classification score profile, which is also amenable to human inspection. Our experimental evaluation shows that ClaSP sets the new state-of-the-art on two benchmark sets of altogether $107$ TS and is also fast and scalable. Open research questions are how to efficiently implement more powerful classifiers for ClaSP, how to approximate it to decrease runtime, how to extend it for multivariate TS, or how to apply it for online segmentation in a streaming setting, like FLOSS and BOCD. We plan to investigate these issues in future work as well as lift the periodicity assumption for ClaSP (compare Section~\ref{sec:background}) to enable it to detect trends in TS beside changes in temporal patterns.

\section{Declarations}\label{sec:conclusion}

The authors of this article have no conflict of interest to declare. 

\bibliography{clasp}

\end{document}